\documentclass[nonacm,acmsmall,screen]{acmart}

\setcopyright{none}

\usepackage{multirow}
\usepackage[title]{appendix}
\usepackage{enumitem}
\usepackage{dsfont}

\theoremstyle{acmplain}

\theoremstyle{acmdefinition}

\newtheorem{definition}{Definition}

\begin{document}

\title[Assessing Predictive Models for Fairness Based on Activity-Space Patterns]{Assessing Predictive Models for Fairness Based on Activity-Space Patterns}

\author{Francesco Lettich}
\correspondingauthor
\email{francesco.lettich@isti.cnr.it}
\affiliation{%
  \department{Institute of Information Science and Technologies}
  \institution{National Research Council}
  \city{Pisa}
  \country{Italy}
}

\author{Mario A. Nascimento}
\correspondingauthor
\email{mario.nascimento@ualberta.ca}
\affiliation{%
  \department{Department of Computing Science}
  \institution{University of Alberta}
  \city{Edmonton}
  \country{Canada}
}

\author{Chiara Pugliese}
\email{chiara.pugliese@iit.cnr.it}
\affiliation{%
  \department{Institute of Informatics and Telematics}
  \institution{National Research Council}
  \city{Pisa}
  \country{Italy}
}

\author{Chiara Renso}
\email{chiara.renso@isti.cnr.it}
\affiliation{%
  \department{Institute of Information Science and Technologies}
  \institution{National Research Council}
  \city{Pisa}
  \country{Italy}
}

\begin{abstract}
Assessing the spatial fairness of predictive models involves establishing whether they are statistically penalizing (favoring) individuals associated with certain geographical locations. Literature on this topic makes the fundamental assumption that each individual is assigned to a single geographical location (e.g., place of residence). 
However, fairness with respect to the set of regions where one regularly spends time, 
i.e., the individual's \emph{activity space}, 
also matters when fairness is considered.
Consequently, we argue that it is necessary to generalize the notion of spatial fairness to also account for such \textit{activity-space patterns}, leading to the novel problem of assessing predictive models for fairness relative to the movements of individuals. 
To deal with this problem, we propose an approach that first associates individuals with geographic regions relevant to their activity spaces, considering multiple spatial partitions with different resolutions and alignments, and then employs a suitable spatial scan statistic to assess whether a predictive model is fair based on activity-space patterns. 
In the experimental evaluation, we study the performance of our approach over thousands of synthetic unfair datasets, showing that it is effective at detecting this new type of unfairness and at retrieving the set of objects treated unfairly, while localization performance exhibits a consistent multi-resolution trade-off.
\end{abstract}



\maketitle

\section{Introduction}
\label{sec: intro}

Fairness in machine learning is a 
well-established, large, and active research area \citep{surveyfairness2024}, that addresses the concern that predictive models, whether due to characteristics in the training data or the training process, can be systematically biased towards (or against) certain subsets of a population, due to improper use of {\emph{protected attributes}}.
A classic example is gender bias in hiring models.
\textit{Spatial fairness} is a recently explored variant of fairness \citep{equitensors21, xie2022fairness, Sacharidis23, saxena2025legally} that makes the assumption that every individual is tied to \textit{one fixed geographical location} (e.g., place of residence) and focuses on whether a model penalizes (or favors) individuals in specific geographic areas more than it would be expected by chance.
In other words, spatial fairness is based on the premise that one’s location can be treated as a grouping attribute for fairness assessment, as it might act as a \textit{proxy} for protected attributes.

For example, a model might indirectly (and possibly incorrectly) correlate, e.g., socioeconomic status, ethnicity, or family structure based on where one lives or works.
Such inferred correlation is the main reason why it is important to \emph{assess} the spatial fairness of a predictive model, as it may indicate the need to \emph{audit} the whole process, from data gathering, cleaning, and processing to decision making. Note that while previous related research (e.g. \citep{Sacharidis23}) has used the term \emph{auditing}, we prefer to use the narrower term \emph{assessment}, as the former suggests investigating compliance to a broader standard, which is not our goal.

Existing research on spatial fairness generally takes two complementary directions: one aims to design predictive models that minimize spatial unfairness from the outset \citep{equitensors21}, while the other focuses on assessing existing predictive models for spatial fairness \citep{Sacharidis23, xie2022fairness}. In this work, we focus on the latter. 

As mentioned before, a key assumption in spatial fairness is that each individual is tied to one single fixed location. In many real-world scenarios, however, an individual habitually spends time in several regions  e.g., their home, workplace, preferred supermarket, or gym. We call this set of regions the individual's \emph{activity space}, and any combination of regions drawn from it an \emph{activity-space pattern}. Since the same pattern can occur in the activity spaces of several individuals, it induces a group, and a predictive model may treat that group differently. Activity-space patterns can therefore act as proxies for protected attributes, in the same way a single location can. This is why the notion of spatial fairness must be generalized beyond the single-fixed-location assumption. The importance of doing so echoes the \emph{neighbourhood effect averaging problem} in health geography \citep{kwan2018neap}: characterizing an individual solely by their residential location fails to capture the multiple contexts they experience through their daily mobility.
 
This leads to the \textit{problem of assessing predictive models for fairness based on activity-space} patterns, and applies to predictive tasks where a model considers, as input, among different types of data, the mobility of individuals, and the model might treat individuals with \textit{certain activity-space patterns} differently.

To help the reader understand the type of fairness introduced in our work, consider the example of a binary classifier that decides loan approvals.
Consider now a group of individuals whose weekly routine regularly takes them to a particular \emph{combination} of areas like the surroundings of a certain place of worship, a discount supermarket in a low-income district, and a specific residential block. Even if such a predictive model (correctly) does not directly take as input one’s protected attributes such as religion, socioeconomic status, or others, assume that these particular individuals are nonetheless approved only $40\%$ of the time compared with 60\% for the remaining population. This $20$-percentage-point gap represents a \textit{disparity in model outcomes} suffered by this group of individuals that is purely defined \emph{by the areas in which they habitually stay to perform their activities}. 
We refer to this combination of areas, which need not be geographically contiguous, as an \emph{unfair activity-space pattern}. 
The underlying problem is referred to as \emph{assessing predictive models for fairness based on activity-space} patterns. We reuse this example throughout the paper to ground the notions we introduce.

Two points about it precisely delineate what our work does and does not do. First, the disparity is measured \emph{only} by analysing the predictive model's outputs, specifically, by assessing the disparity between the outcomes of a specific group of individuals who share a certain activity-space pattern, i.e., a certain combination of areas, and those of the rest of the population. We never need to know the ``correct'' decision for any individual, i.e., we are not interested in a predictive model's classification error, hence no ground-truth labels are required to perform an assessment. This makes our notion of fairness grounded on that of \emph{group fairness} targeting \textit{statistical parity} \citep{dwork2018group} (Section \ref{sec: preliminaries} provides more details on this). 
Secondly, the mere fact that many individuals regularly visit or stay in a, e.g., busy area is not, in itself, evidence of a lack of fairness: places of interest are crowded by design. Hence, what signals unfairness is not the \emph{density} of individuals or movement in some region-combination, but the systematic \emph{difference in model outcomes} for the group of individuals whose activity-space pattern is associated with those region(s) with respect to the rest of the population.
Detecting dense regions is not the problem we address;
rather, we aim to detect region-combinations whose \textit{associated individuals} are treated differently.

There are three dimensions concerning activity-space patterns that existing assessment approaches targeting spatial fairness do not capture:
\begin{description}[style=nextline]
    \item[\textbf{Dimension \#1: Multiple locations/regions.}]
    This is due to the very nature of an individual's movements, i.e., they are bound to be in different regions of the space of interest at different points in time.
    \item[\textbf{Dimension \#2: Frequency of visits to different locations.}]
    It does not suffice to consider that an individual was in given regions, but also how often that was the case.
    \item[\textbf{Dimension \#3: Duration of visits to each location.}]
    Related to the second dimension, it is also relevant how long one was at a given location.
\end{description}
The first one is key to the distinction between spatial fairness and fairness based on activity-space patterns. The other two introduce the previously unaddressed temporal aspect to the \emph{spatial fairness} problem, and allow proper consideration of relevant nuances related to one's activity-space patterns. 
For instance, it may be more meaningful to associate a person with a region they visit regularly for some time, e.g., those containing their home, workplace, supermarket, gym, etc., than those associated with a gas station or a diner they visit rarely and/or briefly.

In summary, this paper makes the following contributions: (i) we formulate the problem of assessing predictive models for fairness based on activity-space patterns, generalizing spatial fairness beyond the single-fixed-location assumption; (ii) we propose an assessment approach that first associates each individual with the regions relevant to their activity space across multiple spatial partitions with different resolutions and alignments -- this is done to mitigate the well-known modifiable areal unit problem (MAUP)\footnote{We highlight that we do not aim to demonstrate the MAUP, which in this work we treat as an established problem; rather, we study how it manifests in our previously unexplored setting.} \citep{openshaw1979million, wong2004modifiable}, and subsequently applies a spatial scan statistic to detect unfair activity-space patterns. We also show that our approach generalizes the spatial fairness assessment from \citet{Sacharidis23}; (iii) we introduce the first protocol to generate synthetic auditable datasets with injected unfairness based on activity-space patterns; and (iv) we evaluate our assessment approach over thousands of synthetic auditable datasets to empirically characterize its detection, retrieval, and localization performance across partitions with different spatial resolutions and alignments.

The remainder of the paper is organized as follows. Section \ref{sec: related} discusses related work, noting that, to the best of our knowledge, none of them can address the problem we propose. Based on the motivation and identification of the research gaps introduced above, Section~\ref{sec: preliminaries} establishes the necessary background for our proposed approach to be presented.
Section~\ref{sec:approach} presents the main contribution of this paper, namely,  an approach that first associates individuals with the geographical regions that are relevant to their activity-space patterns, and then uses a suitable spatial scan statistic to find out if there are individuals sharing certain activity-space patterns that are treated unfairly.
In the experimental evaluation (Sections \ref{sec:experimental setting} and \ref{sec:experimental evaluation}), we first explain why thousands of unfair synthetic datasets are needed to evaluate our approach, and how to generate them, noting that our work is the first to propose a generation protocol tailored to unfairness based on activity-space patterns. We then use these datasets to study the effectiveness of the proposed approach with respect to several parameters. We show that it can effectively detect the presence of unfairness, retrieve the set of objects treated unfairly, and, depending on how the space is partitioned, localize it satisfactorily.
Finally, we outline a practical exploratory strategy for real-world assessments and illustrate its application through an example (Section~\ref{sec: exploratory strategy}).
\section{Related work}
\label{sec: related}

Fairness in machine learning has been an active research area concerned with the risk that predictive models may produce biased outcomes for certain subsets of a population. The literature distinguishes between \textit{individual fairness} \citep{dwork2018individual}, which requires that similar individuals be treated similarly, and \textit{group fairness} \citep{dwork2018group}, which requires that groups sharing a common protected attribute receive statistically comparable treatment. Throughout this work, we use the term fairness to refer to group fairness unless otherwise stated. In Appendix \ref{app: supplemental group fairness stat parity} we provide a formal definition of group fairness based on statistical parity. 
To avoid ambiguity, from here on we use the term \emph{fairness} to denote the criterion being assessed and \emph{bias} to denote the systematic disparity in model outcomes that results from its violation; a violation of group fairness thus constitutes bias against (or in favor of) the affected group.

\textit{Spatial fairness} is a variant of fairness where location is considered as the protected attribute, observing that it can act as a \textit{proxy} for other protected attributes (e.g., race or income).
In the following, we focus on works that assess the spatial (group) fairness of existing predictive models.
\citet{xie2022fairness} propose a measure that assesses spatial fairness by evaluating how a classifier's performance varies across multiple geographic subdivisions. To mitigate the Modifiable Areal Unit Problem (MAUP) \citep{openshaw1979million, wong2004modifiable}, which is related to the fact that spatial analysis results can change depending on the scale and boundaries of the chosen partitions, the measure repeats this evaluation over many different partitions and summarizes the results as the average variance of the classifier's performance across them.
However, this measure is sensitive to uneven spatial distributions of labeled data and requires access to ground truth labels, as its assessment is based on predictive performance, in this specific case, on classification error. This differs from the statistical-parity perspective adopted in our work and in others \citep{Sacharidis23, saxena2025legally}, which focuses on disparities in model outcomes across groups and does not require ground-truth labels.
Furthermore, it does not account for the fact that poor predictive performance in a region can be due to causes unrelated to fairness, e.g., limited local training data, noisy labels, or distribution shift. Finally, predictive performance and fairness can actually be competing objectives, and sacrificing the former might be required \citep{dwork2018group, friedler2019comparative, Shaham22}.

\citet{Sacharidis23} propose a statistically grounded approach to assess binary classifiers for spatial fairness.
Their approach uses the Bernoulli-based spatial scan statistic from \citet{kulldorff1997spatial}, previously used to detect hotspots of diseases, to instead find out the regions in which the proportion of positive labels deviates significantly from the proportion outside. To mitigate the MAUP, their approach also supports using the statistic considering multiple different tessellations.
The authors did not quantitatively evaluate the detection and localization performance of their approach; while this can be justified by the fact that they apply the spatial scan statistic of \citet{kulldorff1997spatial} as is, it remains unclear to what extent their approach is MAUP-resistant, and they did not provide principled guidelines on how to choose the tessellation resolutions.

A subsequent work \citep{saxena2025legally} builds on \citep{Sacharidis23} with a key observation: statistical unfairness can actually have legitimate explanations. It then assumes that information concerning the protected and non-protected attributes of individuals are known and, for any cell in any tessellation over the area of interest, their approach derives two vectors: one related to the protected attribute and the other to the non-protected ones of the cell's individuals. They then identify pairs of cells that are similar in the non-protected attributes but dissimilar in the protected one, and identify the pairs whose outcome distributions differ substantially using a modified version of the Bernoulli-based spatial scan statistics, which compares pairs of regions rather than a region against the rest of the space.
\citep{saxena2025legally}'s approach relies on task-dependent similarity metrics that may require deep and ad-hoc domain knowledge to define. Moreover, while the approach is claimed to be MAUP-resistant, it still remains vulnerable to adversarial repartitioning that dilutes unfair outcomes by merging affected areas with neighboring fair ones; furthermore, no principled guideline on how to choose the tessellation resolutions was provided. Finally, the authors did not measure the detection and localization performance of their approach, which is standard practice when proposing a new spatial scan statistic \citep{read2011measuring, kulldorff2009scan, jung2010spatial}.

To the best of our knowledge, our work differs from previous literature on spatial fairness as it considers the problem of assessing predictive models for fairness based on the activity-space patterns of individuals, which requires going beyond the basic assumption that each individual is associated with a single fixed location.
\section{Preliminaries and problem definition}
\label{sec: preliminaries}

In this section, we review some fundamental concepts that will be used throughout the rest of the paper, and introduce the problem addressed in this work.
We denote the set of moving objects by $MO$ and the set of their trajectories by $\mathcal{T}$. 
First, we describe the structure that a dataset should have in order to allow assessing a predictive model for fairness based on activity-space patterns.

\begin{definition}[Auditable Dataset]
\label{def:auditable dataset}

Each element in an \emph{auditable dataset} $D$ is a triplet $(mo, T_{mo}, pred_{mo})$, where $mo \in MO$ is a moving object, $T_{mo} \in \mathcal{T}$ is the trajectory describing $mo$'s movements, and $pred_{mo}$ is a value predicted by the model for $mo$.
\end{definition}

Before considering the notion of spatial fairness, we first introduce the notion of geographic zoning, which is necessary for identifying \textit{regions} where unfairness by a predictive model might occur.

\begin{definition}[Geographic zoning]
\label{def: geographic zoning}

Consider a geographic area of interest that contains all trajectories in $\mathcal{T}$. Suppose that within such an area we define a set of polygonal cells $GZ$ that partitions the area into disjoint regions, e.g., a uniform grid.
We refer to any such set $GZ$ as a \emph{geographic zoning} of the area of interest.
\end{definition}

We now introduce the notion of \textit{spatial fairness} \citep{Sacharidis23}. From a technical standpoint, it can be viewed as an instance of group fairness targeting statistical parity where the location an object is associated with is the protected attribute.  \citep{dwork2018group, friedler2019comparative} (Appendix \ref{app: supplemental group fairness stat parity} provides a formal definition). Assessing it therefore amounts to determining whether the objects associated with a given region receive model outcomes that differ systematically from those of the rest.

Our problem generalizes this beyond a single location. Recall from Section \ref{sec: intro} that an object's habitual stays define its \emph{activity space}, and that any combination of the corresponding cells is an \emph{activity-space pattern}. Furthermore, activity-space patterns are shared, in general, by several objects. An \emph{unfair activity-space pattern} is an activity-space pattern whose associated objects receive model outcomes whose distribution differs significantly from that of the remaining objects; those objects form the group that is treated unfairly. For brevity, and consistently with the spatial scan statistics literature, we will refer to an unfair activity-space pattern also as a \textit{hotspot of unfairness}, or simply a \textit{hotspot}.

From these considerations, two consequences follow that help dissipate two common misreadings of this notion. 
First, a hotspot need not be spatially contiguous: it is a \textit{set} of cells bound together by their association with the same moving objects, not by adjacency, so it may consist of several separate regions. Second, the evidence relevant to the fairness assessment comes from disparities in model outcomes, not from mobility or presence density: a region-combination simply visited by many objects does not by itself provide evidence of unfair treatment, whereas a region-combination whose associated objects exhibit a statistically significant disparity in model outcomes may do so. Note that this is an instance of group fairness targeting statistical parity, where the ``group'' is induced by a subset of cells occurring in the activity-space patterns of its members, and the disparity is a difference between the outcome distribution \textit{inside} the group and \textit{outside} it.

Assessing either for spatial fairness or fairness based on activity-space patterns also requires dealing with the Modifiable Areal Unit Problem (MAUP) \citep{wong2004modifiable, xie2022fairness, Sacharidis23}, i.e., the spatial analysis results can change with the scale and zoning/boundaries of the chosen spatial partitions (Appendix \ref{app: supplemental MAUP example} provides an illustrative example).
Consequently, we use several different geographic zonings $\mathcal{GZ} = \{GZ_1, \ldots, GZ_q\}$ to mitigate the MAUP; let us focus on a single geographic zoning $GZ$.
Establishing which subset of cells in $GZ$ should be assigned to each moving object, based on their spatiotemporal behavior and/or domain-oriented spatial and temporal thresholds, is not a trivial task, mainly due to the three dimensions posed in Section \ref{sec: intro}.

Furthermore, the MAUP has an additional implication in the context of fairness based on activity-space patterns, as it needs to be considered under the light of a moving entity, unlike in the case of conventional spatial fairness, which assumes that each entity is associated with a single fixed location.
Indeed, a moving object may be associated with multiple geographically distinct regions, and changing the resolution or boundaries of the geographic zoning may therefore alter the combination of regions associated with it and, consequently, the groups induced by activity-space patterns. 

This motivates the explicit formalization of how each moving object is associated with the cells of a given geographic zoning.

\begin{definition}[Moving object-to-cell association]
\label{def: moving object to cell association}
Given a set of moving objects $MO$, their trajectories $\mathcal{T}$, and a geographic zoning $GZ \in \mathcal{GZ}$, we define an \emph{association function} $asc: MO \rightarrow 2^{|GZ|}$ that maps each moving object $mo \in MO$ to a subset of cells in $GZ$ according to the characteristics of its trajectory $T_{mo} \in \mathcal{T}$.
\end{definition}

How to instantiate the function $asc$ will be discussed while presenting our approach in Section \ref{sec:approach}. We are now ready to state the problem addressed in this work.\\

\noindent \textbf{Problem Statement.} Given an auditable dataset $D$, a set of geographic zonings $\mathcal{GZ}$ over an area of interest, and a moving object-to-cell association function $asc$, our goal is to identify whether there are subsets of cells in any $GZ \in \mathcal{GZ}$ whose associated moving objects have model outcomes that differ systematically relative to the other moving objects\footnote{This means that for the moving objects associated with any subset of cells of a given geographic zoning, the model’s predicted values deviate significantly from the others, as established by a suitable hypothesis test.}.
\section{The proposed approach }
\label{sec:approach}

Next, we present our proposed approach to assess predictive models for fairness based on activity-space patterns. The approach is structured in five distinct steps.
First, it \textit{segments trajectories into stop and move segments} (Section \ref{sec: traj segmentation}). Stop segments correspond to parts of a trajectory where an object remained around a certain location for some time, while move segments capture transitions between stops. 
Subsequently, our approach \textit{superimposes} several \textit{geographic zonings}, i.e., spatial partitions, using different resolutions and alignments, over the area containing the segments (Section \ref{sec: space partitioning}). This is done to mitigate the Modifiable Areal Unit Problem (MAUP) \citep{wong2004modifiable, xie2022fairness, Sacharidis23}, as considering multiple partitions reduces this risk.

The approach then repeats the following two steps for each partition. It first performs \textit{object-to-cells mapping} (Section \ref{sec: users to cells mapping}), i.e., it maps each moving object to the subset of cells it is regularly associated with, whose subsets represent the activity-space patterns associated with that object.
Subsequently, it uses this mapping to \textit{compute the subsets of cells that must undergo hypothesis testing} (Section \ref{sec: determining cells audit}). After this, the approach applies a suitable \textit{spatial scan statistic on such subsets}, to assess whether there is statistically significant evidence of unfairness based on activity-space patterns and identify the subsets of cells associated with such evidence; and \textit{returns the evidence} (if any) allowing the user to further investigate the potential bias (Section \ref{sec: hypotesis test}).

Finally, we conclude this section by showing how our assessment approach can reduce to the one from \citet{Sacharidis23} under the spatial fairness assessment scenario (Section \ref{sec: reduction to spatial fairness}).

\subsection{Trajectory segmentation}
\label{sec: traj segmentation}

Our approach begins by instantiating the \textit{object-to-cells association} function $asc$ introduced in Definition \ref{def: moving object to cell association}, thus starting to address the three dimensions posed in Section \ref{sec: intro}. Recall that this function must ultimately map each moving object to a subset of geographic areas that meaningfully characterize its activity space.

As a first step toward identifying such cells, our approach first performs an operation known as \textit{trajectory segmentation}, using the well-known \textit{stop-and-move} criterion \citep{spaccapietra2008conceptual}. 
A stop segment represents a specific part of a trajectory during which the associated moving object \textit{stayed} in a \textit{specific area} for \textit{some time}. The move segments are simply parts of a trajectory that are not stop segments.
In this work, we focus specifically on the stop segments and use the algorithm from \citep{hariharan2004project} to find them.
We adopt a person-based perspective in which we aim to identify the geographic locations where an individual habitually stays, that is, the stop locations that consistently structure their routine and hence \textit{characterize their activity space}. Typical examples include home, work, school, gym, preferred supermarkets or restaurants, and more.

Formally, we represent each stop segment as a quadruple $st = (mo, l, t_{start}, t_{end})$, where $mo$ is the identifier of the moving object, or individual, $l$ is the location of the centroid of the stop segment and $t_{start}, t_{end}$ its starting and ending instants of the stop, respectively. We finally denote the set of all detected stop segments as $STOP = \{st_1, \ldots, st_m\}$, where $m$ is the total number of stops detected across all moving objects’ trajectories in the auditable dataset $D$.

\subsection{Superimposing multiple geographic zonings}
\label{sec: space partitioning}

Following the terminology introduced in Definition \ref{def: geographic zoning} and considering the modifiable areal unit problem (MAUP), this step superimposes $q$ different geographic zonings $\mathcal{GZ} = \{GZ_1, \ldots, GZ_q\}$ over the geographic area containing the locations associated with the stop segments.
A natural question is then: how to select an appropriate number of zonings $q$?
In practice, this requires deciding how many different resolutions and alignments one wants to use to mitigate the MAUP.

Part of this choice concerns the \textit{coarsest} grid resolution, and is informed by some spatial scan statistics literature \citep{kulldorff2001prospective} which suggests constraining the size of the cells to contain at most 50\% of the total population. This excludes zonings that are, e.g., too coarse.
The remaining choices concern the \textit{finest} and \textit{intermediate} grid resolutions. These depend on the size and structure of the geographic area under study, the spatial scales at which meaningful activity-space patterns may arise, and the available computational budget. In general, the finest resolution should reflect the smallest relevant geographic unit in the considered area, such as a city block in an urban setting. The intermediate resolutions should cover the range between the finest and coarsest resolutions, so that hotspots (i.e., unfair activity-space patterns) can be detected at multiple spatial scales.
As for the alignments, one should use a fixed number of shifts for each chosen resolution to better capture hotspot boundaries.
Section \ref{sec: exp eval parameters} provides a concrete instantiation of these choices for our study area, from $50$-meter city-block cells up to $1000$-meter cells.

\subsection{Object-to-cells mapping}
\label{sec: users to cells mapping}

This step takes as input the stop segments computed in the previous trajectory segmentation step and a single geographic zoning $GZ \in \mathcal{GZ}$, and addresses the problem of associating each moving object with the set of geographical regions (cells) in $GZ$ that characterize its activity space. This is again connected to dealing with the three dimensions from Section \ref{sec: intro}, as individuals may habitually stay in multiple regions. Depending on how frequently and for how long people visit such regions, some of these regions are more relevant than others.

Our approach performs this mapping by analyzing where and when the stop segments occur within the cells of the geographic zoning. Let $\mathit{STOP}_{mo} \subseteq \mathit{STOP}$ be the set of stop segments for the moving object $mo$. We first perform a spatial join between the centroids of the stop segments and the cells in $GZ$ based on the \textit{intersection} predicate: 
$
STOP_{mo}^{ann} = \mathit{STOP}_{mo} \; \underset{\mathrm{intersect}}{\overset{\mathrm{spatial}}{\Join}} \; GZ.
$
This generates a set of annotated stops, $\mathit{STOP}_{mo}^{ann}$, where each stop segment becomes a quintuple $(mo, l, cell, t_{start}, t_{end})$, and allows us to identify the subset of cells in $GZ$ that contain at least one of $mo$'s stop segments. We denote this subset as $GZ_{mo}$.

Notice that not all cells touched by a moving object’s stop segments are equally meaningful. For example, one cell may contain a place of worship, 
with repeated long stays, while another may include only a single occasional stay, e.g., the person visited a new restaurant to try it out; clearly, these regions carry different relevance.
To identify the most relevant regions, we apply the following criterion: for every cell in $GZ_{mo}$, we count the \textit{number} of \textit{distinct days} temporally intersecting one or more of the stop segments of $mo$ in that cell. Intuitively, the more distinct days in which a cell occurs in a person's life, the more that cell is relevant to the person's activity space.
Note that while in this paper we adopt ``distinct days'' as a unit of interest, other units can be used without prejudice to the proposed approach.

This criterion ultimately induces a ranking over the cells in $GZ_{mo}$ and, together with the preceding trajectory-segmentation step, addresses the aforementioned three dimensions characterizing our problem.
We then select the top-$i$ cells from $GZ_{mo}$ and define them as the \textit{cellset} of $mo$.
This cellset represents \textit{mo}’s activity space, while its subsets represent the activity-space patterns associated with \textit{mo}.
We denote this set by $GZ_{mo}^{i}$ and use it in the subsequent steps of the assessment process. The parameter $i$ acts as a cut-off: it determines how many cells are deemed relevant for each moving object. In practice, $i$ should reflect the typical number of regions relevant to people’s activity spaces, with some margin for variability (e.g., $i=8$). 
Going back to the example of the loan applicants from Section \ref{sec: intro}, this criterion would rank highly the cells covering the place of worship, the discount supermarket, and the residential block -- since each of these cells recurs on many distinct days -- and retain them within the object's cellset.
This last operation terminates the instantiation of the \textit{object-to-cells association} function $asc$ introduced in Definition \ref{def: moving object to cell association}. Finally, recall that the sequence of operations just described is repeated for every zoning $GZ \in \mathcal{GZ}$.

\subsection{Determining the activity-space patterns to assess}
\label{sec: determining cells audit}

For each geographic zoning $GZ \in \mathcal{GZ}$, this step takes as input the cellsets computed for the moving objects in the previous step, and computes the subsets of cells that must undergo hypothesis testing. 
More precisely, since subsets of a moving object's cellset represent its activity-space patterns, we aim to find the subsets of cells that are contained in at least one moving object's cellset and therefore correspond to activity-space patterns associated with at least one object.
The main challenge in solving this problem lies in the size of the search space. Each moving object’s cellset is made up of up to $i$ cells, and cellsets can intersect in many different ways; thus, if $GZ$ is the geographic zoning currently considered, then $|GZ|$ is the number of cells in this particular zoning, and the search space has exponential size $O(2^{|GZ|})$. Consequently, efficiently exploring this space, both in terms of memory and computation, is crucial.

To address this problem, we use techniques from the well-known \textit{frequent itemset mining} (FIM) literature \citep{agrawal1996fast, zaki2000scalable}, effectively casting the problem faced in this step as a variant of FIM. 
Accordingly, we first provide a brief introduction to the FIM problem, adopting the terminology from one of the seminal works in the literature \citep{zaki2000scalable}. 
In FIM, one is given a typically large database of transactions where each transaction contains a set of items. The goal is to discover the so-called \textit{frequent itemsets}, i.e., subsets of items whose \textit{support}, i.e., number of transactions they occur in, is above a given threshold. For example, in market basket analysis, each transaction represents the products bought by a customer, and finding the frequent itemsets across many transactions means finding the sets of products that customers often buy together.
Going back to our problem, we can cast it as an FIM variant as follows:
\begin{itemize}[leftmargin=1.2em]
\item Each moving object's cellset, $GZ^i_{mo}$, can be seen as a transaction uniquely identified by the object's identifier; 
\item Each cell can be seen as an item that appears in at least one cellset (transaction);
\item The \textit{support} of a cell is equal to the number of cellsets in which it occurs;
\item From the FIM literature (e.g., lemma 5 from \citep{zaki2000scalable}), itemsets (subset of cells) can be \textit{joined} into larger itemsets by taking their set-union; and the IDs of the transactions associated with a joined itemset are obtained by set-intersecting the sets of IDs associated with the two itemsets being joined; 
\item Therefore, the \textit{support} of a \textit{subset of cells} is equal to the number of moving objects whose cellsets contain the subset. Thus, each such subset represents an activity-space pattern and its support identifies the size of the group induced by that pattern.
\end{itemize}
For example, suppose three objects have cellsets $\{A,B\}$, $\{A,B,C\}$, and $\{A,D\}$. Then, cell A has support 3, the subset $\{A,B\}$ has support 2, and $\{A,B,C\}$ has support 1. Moreover, $\{A,B\}$ represents an activity-space pattern shared by the first two objects and induces a group of two objects for fairness assessment.

Thus, the problem is to identify all subsets of cells in $GZ$ with support at least one, i.e., all subsets associated with at least one object, and denote the resulting set by $\mathit{CANDIDATES}_{GZ}$. We intentionally retain all such subsets so as not to exclude activity-space patterns associated with only a small number of objects, while subsets with zero support are untestable and thus discarded.
The algorithm implementing this step has the same key characteristics as the Eclat candidate generation algorithm from \citep{zaki2000scalable}. Appendix \ref{app: supplemental complexity cand gen} provides more details on the algorithm and this step's complexity.

\subsection{Hypothesis testing}
\label{sec: hypotesis test}

This step returns to considering all the zonings in $\mathcal{GZ}$, and aims to determine whether there is statistically significant evidence of a disparity in model outcomes for the group induced by any activity-space pattern represented by a subset of cells in $\mathit{CANDIDATES}_{\mathcal{GZ}} = \{ \mathit{CANDIDATES}_{GZ_1} \cup \ldots \cup \mathit{CANDIDATES}_{GZ_q} \}$, according to the predicted values available in the auditable dataset $D$. 

 Before introducing this step's formal stages, we provide the intuition behind it: we measure \textit{how surprising} the observed disparity in predicted values is for the objects associated with the most extreme candidate subset of cells in $\mathit{CANDIDATES}_{\mathcal{GZ}}$, by comparing it against many datasets in which the predicted values have been randomly reshuffled among the objects. Under such reshuffling, no group of objects induced by an activity-space pattern should stand out; if the real data produce a more extreme disparity than almost all of the reshuffled datasets, we take this as evidence of unfairness.

Let us denote the set of predicted values in $D$ as $PRED$.
We resort to a well-established family of spatial scan statistics  \citep{kulldorff1997spatial, kulldorff2009scan, jung2010spatial}.
While the specific spatial scan statistic required depends on the predictive model being assessed, the stages of the high-level testing methodology are as follows:\\

\noindent \textbf{Stage 1: State the null hypothesis $\mathbf{H_0}$ and alternative $\mathbf{H_1}$}. The null hypothesis $H_0$ posits that there is no disparity in model outcomes associated with any candidate activity-space pattern, i.e., all moving objects share the same global distribution for the predicted values.
The alternative $H_1$ posits that there is at least one activity-space pattern $c \in \mathit{CANDIDATES}_{\mathcal{GZ}}$ whose induced group of moving objects has a distribution of predicted values that differs from that of the remaining objects.

\vspace{0.2em}
\noindent \textbf{Stage 2: Choose the spatial scan statistic appropriate for the model being assessed.} A test statistic $T$ is a function that maps the observed data (predicted values) to a scalar, and is used to decide whether to reject the null hypothesis $H_0$ or not.    
The family of spatial scan statistics from Kulldorff et al. that we consider in this work employs the \textit{maximum likelihood ratio} (MLR) as $T$. Each of these statistics instantiates the MLR according to the parametric probability model appropriate for the predictive model being assessed, e.g., a Bernoulli-based statistic for binary classifiers \citep{kulldorff1997spatial}, a multinomial-based statistic for multi-class classifiers \citep{jung2010spatial}, and a Normal-based statistic for regressors \citep{kulldorff2009scan}.
    
In general, the MLR requires maximizing two likelihood functions $L_0$ and $L_1$ that formally represent $H_0$ and $H_1$, respectively, and are defined using the chosen parametric probability model. They express how plausible the observed data is under each hypothesis when using specific parameter(s) for the probability model. Maximizing $L_0$ and $L_1$ yields $L_0^{max}$ and $L_1^{max}$, which lead to the MLR: 
$$T = \frac{L_1^{max}}{L_0^{max}}$$ 
The larger $T$, the more the predicted values favor $H_1$ over $H_0$, i.e., the moving objects in the group induced by some $c \in \mathit{CANDIDATES}_{\mathcal{GZ}}$ exhibit a disparity in model outcomes beyond what would be expected by chance.
Note that a single value of $T$ does not determine whether to reject $H_0$, as we need to establish how surprising that value is under $H_0$. We also need the distribution of $T$ under the assumption that $H_0$ is true. In spatial scan statistics, this is approximated through Monte Carlo simulations, obtaining $T$'s \textit{null distribution} (see Stage 6). We then compute $T$ on the actual predicted values $\mathit{PRED}$, and see where it falls in the null distribution to decide whether to reject $H_0$ (Stages 7 and 8).

\vspace{0.2em}
\noindent \textbf{Stage 3: Formalize $\mathbf{H_0}$}. $H_0$ is formalized with the likelihood function $L_0(\theta; \mathit{VALUES})$, which uses the analytical form of the chosen parametric probability model to measure how plausible the data in $\mathit{VALUES}$ are under the assumption that there is a single global distribution over the predicted values. The goal is to find the specific $\theta$ that maximizes $L_0$: 
$$L_0^{max}(\mathit{VALUES}) = \max_{\theta} L_0(\theta; \mathit{VALUES}).$$
    
\vspace{0.2em}
\noindent \textbf{Stage 4: Formalize $\mathbf{H_1}$.} We still use the same probability model, but use it twice when defining $L_1$: one for the moving objects in the group induced by some activity-space pattern $c \in \mathit{CANDIDATES}_{\mathcal{GZ}}$, with input parameters $\theta_{in}$, and another for the other objects, with input parameters $\theta_{out}$. Hence, we get $L_1(\theta_{in}, \theta_{out}, c, VALUES)$.
Analogously to $L_0$, the goal is to maximize $L_1$, but with an additional step: we must first maximize $L_1$ separately for each $c \in \mathit{CANDIDATES}_{\mathcal{GZ}}$ 
$$L_1^{max}(c, \mathit{VALUES}) = \max_{\theta_{in}^c, \theta_{out}^c} L_1(\theta_{in}^c, \theta_{out}^c; c, \mathit{VALUES}),$$
and finally, find out the $c$ that yields the largest value
\begin{equation*}
L_1^{max}(\mathit{CANDIDATES}_{\mathcal{GZ}}, \mathit{VALUES}) =
\max_{c \in \mathit{CANDIDATES}_{\mathcal{GZ}}} L_1^{max}(c, \mathit{VALUES}).
\end{equation*}
In essence, $H_1$ is a composite alternative stating that there exists at least one subset $c \in \mathit{CANDIDATES}_{\mathcal{GZ}}$ for which $\theta_{in} \neq \theta_{out}$.

\vspace{0.2em}
\noindent \textbf{Stage 5: Choose a statistical significance level $\mathbf{\alpha}$}. It represents the maximum probability of rejecting $H_0$ when it is actually true.

\vspace{0.2em}
\noindent \textbf{Stage 6: Compute an approximated distribution of the test statistic under $\mathbf{H_0}$ via Monte Carlo Simulations.} 
Following standard practice in spatial scan statistics, we derive an approximation of $T$'s distribution from the predicted values in \textit{PRED} under the assumption that $H_0$ is true, which allows us to determine how \textit{surprising} the observed value of $T$ is under $H_0$.
More specifically, we conduct $n$ Monte Carlo simulations: in each simulation $b \in \{1, \ldots, n\}$, we generate a synthetic dataset of predicted values $\mathit{PRED}_b$ by randomly shuffling the values in \textit{PRED}.
The approximated null distribution of $T$ is then obtained by computing for each simulation $b$ the MLR:
$$T_b = \frac{L_1^{max}(\mathit{CANDIDATES}_{\mathcal{GZ}}, \mathit{PRED}_b)}{L_0^{max}(\mathit{PRED})}.$$
Concerning $L_0^{max}(\mathit{PRED})$, recall that $H_0$ postulates that there is just a single global probability distribution over all the moving objects' predicted values. Thus, shuffling the values in $\mathit{PRED}$ does not change such a distribution, hence $L_0^{max}(\mathit{PRED})$ can be computed once and shared across all the simulations. However, $L_1^{max}(\mathit{CANDIDATES}_{\mathcal{GZ}}, \mathit{PRED}_b)$ must be computed for each simulation, as it depends on the shuffled values and the partition $GZ \in \mathcal{GZ}$ being considered.
Ultimately, $T^{H_0} = \{T_1, \ldots, T_n\}$ provides the approximated null distribution.
    
\vspace{0.2em}
\noindent \textbf{Stage 7: Compute the test statistic over the original predicted values in \textit{\textbf{PRED}}.} From \textit{PRED} and the set of subsets of cells to be tested in $\mathit{CANDIDATES}_{\mathcal{GZ}}$, compute the scalar:
$$T_{obs} = \frac{L_1^{max}(\mathit{CANDIDATES}_{\mathcal{GZ}}, \mathit{PRED})}{L_0^{max}(\mathit{PRED})}.$$

\vspace{0.2em}
\noindent \textbf{Stage 8: Determine whether to reject $\mathbf{H_0}$.}
With $T^{H_0}$ and $T_{obs}$, we can estimate the probability under $H_0$ of observing a MLR \textit{at least as extreme as} $T_{obs}$, i.e., $\hat p = \Pr(T \ge T_{obs}\mid H_0)$. 
Subsequently, we compare $\hat p$ to the significance level $\alpha$ and reject $H_0$ if $\hat p \le \alpha$. Intuitively, the larger (more extreme) $T_{obs}$ is compared to the top values in the sorted $T^{H_0}$, the more surprising the observed disparity is under $H_0$, and therefore the stronger the statistical evidence against $H_0$. Concretely, we first determine the $p$-value of $T_{obs}$ as its rank in $T^{H_0}$:
$$\hat{p} = \frac{1 + \sum_{T_b \in T^{H_0}} \mathds{1}(T_b \ge T_{obs})}{n + 1},$$
where $\mathds{1}(\cdot)$ represents the indicator function and $n$ the number of simulations. Then, if $\hat{p} \le \alpha$ we \textit{reject} $H_0$ in favor of $H_1$.

\vspace{0.2em}
\noindent \textbf{Stage 9: preparing the evidence for further analysis}. If $H_0$ is rejected, then there is at least one activity-space pattern whose induced group of objects exhibits statistically significant evidence of a disparity in model outcomes, and others might exist. Hence, for each $c \in \mathit{CANDIDATES}_{\mathcal{GZ}}$ we consider $T_c = \frac{L_1^{max}(c, \mathit{PRED})}{L_0^{max}(\mathit{PRED})}$ -- observe that $L_1^{max}(c, \mathit{PRED})$ has already been computed for $L_1^{max}(\mathit{CANDIDATES}_{\mathcal{GZ}}, \mathit{PRED})$ in Stage 7 -- and find its rank in $T^{H_0}$ to verify if it is less than or equal to $\alpha$: all the subsets (activity-space patterns) that satisfy this condition are stored in $\mathit{UNFAIR}_{\mathcal{GZ}}$.\\

Appendix \ref{app: supplemental complexity hyp test} provides additional considerations regarding the complexity of this step, while Appendix \ref{app: supplemental complexity approach} provides some considerations about the overall complexity of our approach.

\subsection{Reduction to the standard spatial fairness assessment case}
\label{sec: reduction to spatial fairness}

It is worth noting that our approach is also capable of evaluating the spatial fairness of a predictive model under the scenario considered in \citep{Sacharidis23}; thus, being more generic. Recall that in that more constrained scenario, each object is associated with a single point. Therefore, the \textit{trajectory segmentation} step is unnecessary and can be skipped. 
Our approach still executes the step that \textit{superimposes multiple geographic zonings} with different resolutions and alignments, over the area containing the points. 
Next, for each geographic zoning, the \textit{object-to-cells} mapping step reduces to a spatial join (using the intersection predicate) between the points and the cells of the zoning, while the step that \textit{determines} the subsets of cells that \textit{must undergo hypothesis testing} also simplifies, since it only needs to identify the cells that contain at least one point. Finally, the hypothesis testing recipe remains unchanged.
\section{Experimental setting}
\label{sec:experimental setting}

In this work, we instantiate the experimental evaluation for the case of assessing a binary classifier for fairness based on activity-space patterns. 
Section \ref{sec: exp generation synthetic movement data} describes why we need to generate synthetic unfair auditable datasets, and how we generate synthetic movement data through a simulator. Section \ref{sec: exp eval unfairness injection} introduces the protocol we use to generate thousands of unfair auditable datasets from said movement data.
Section \ref{sec: evaluation methodology} presents the measures used to evaluate our assessment approach. Finally, Section \ref{sec: exp eval parameters} details the parameters used to configure the approach for the experimental evaluation.
 
Although our experimental evaluation focuses on assessing a binary classifier, we highlight that our approach is not limited to this case. 
In order to consider other types of predictive models, it would require (1) using a spatial scan statistic in the hypothesis testing in Section \ref{sec: hypotesis test} that is appropriate for the model being assessed; (2) adjusting the protocol so that unfairness is injected considering a type of distribution that is appropriate for the values outputted by the model; and (3) repeating the empirical evaluation on the newly generated unfair datasets.
In essence, our approach remains the same, what changes are the spatial scan statistics and the protocol to generate unfair datasets. 

The source code behind this work is available at: \url{https://github.com/Fr4nz83/AuditMovementFairness}.
The datasets used in this work and the intermediate data generated by the proposed approach are available at: \url{https://figshare.com/s/93b04d0a6128d3e7ca32}.

\subsection{Generation of synthetic movement data}
\label{sec: exp generation synthetic movement data}

To the best of our knowledge, there are no real auditable datasets that meet the needs of our evaluation. As established in the spatial scan statistics literature \citep{read2011measuring, kulldorff2009scan, jung2010spatial}, estimating the detection and localization performance of an approach that uses a spatial scan statistic requires using measures\footnote{These measures are detailed in Section \ref{sec: evaluation methodology}.}
that evaluate the approach over large collections of datasets in which the presence and location of unfairness are known by construction.
We therefore need to generate numerous synthetic, auditable datasets. 
We do this by first generating synthetic movement data, which we then use as the basis to generate unfair auditable datasets. The idea is to keep the movement data fixed, while unfairness is injected via the objects' predicted values.

\begin{figure}[h]
\centering
\includegraphics[width=0.5\linewidth]{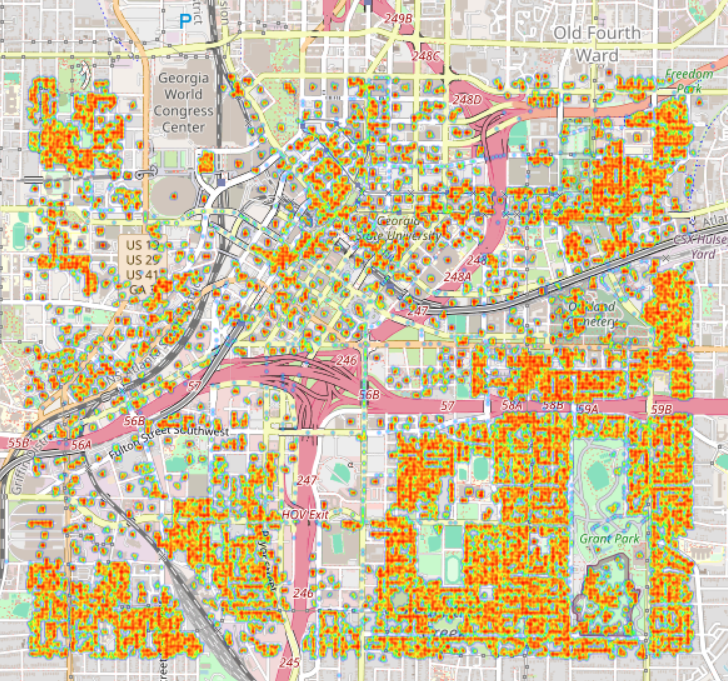}
\caption{Heatmap of a million trajectory samples  where the red areas denote the locations where the simulated individuals tend to stay.}
\Description{Street map of the simulated study area in Atlanta overlaid with a heatmap based on one million trajectory samples. Orange and red clusters indicate locations where simulated individuals tend to stay. The samples are unevenly distributed across the study area, with multiple areas of concentrated activity separated by areas with lower densities.}
\label{fig:simulator trajs heatmap}
\end{figure}

We generate the movement data using the \textit{Patterns of Life} simulator from \citet{zufle2023urban}. We simulate 100,000 individuals living in the city of Atlanta, Georgia, USA. Their movements follow well-defined behavioral patterns that reflect basic human needs, such as commuting between home, work, supermarket, restaurant, and more. Figure \ref{fig:simulator trajs heatmap} shows a heatmap based on a random sample of 1 million trajectory points drawn from the original 720,199,539, in which we notice their uneven spatial distribution. The sampling rate of the trajectories is 2 minutes. The temporal interval spanned by the trajectories is 10 days. The bounding box enclosing the trajectory samples has a width of 4.69 kilometers and a height of 4.83 kilometers.
From here on, we refer to the simulated individuals as objects.

The simulator generates a very large amount of data, hence we apply the trajectory compression algorithm outlined in \citep{zheng2015} and implemented in the scikit mobility library from \citet{pappalardo2022scikit}.
Briefly, for each object, samples are first ordered by timestamp. Starting from the first uncompressed sample, the algorithm uses it as an anchor point and collects the maximal consecutive sequence of subsequent samples whose distance from the anchor is at most a given \textit{radius}. This sequence is replaced by a single sample whose spatial coordinates are the component-wise median of the coordinates in the sequence and whose timestamp is the timestamp of the anchor sample. The procedure then continues from the first sample outside the radius and is repeated until the trajectory has been fully processed.
In our experiments, we set the above radius to one meter in order to remove very short movements, bringing the final number of trajectory points to 19,815,853.

\subsection{Generating unfair auditable datasets: A protocol for injecting unfairness based on activity-space patterns}
\label{sec: exp eval unfairness injection}

Next, we generate several groups of synthetic unfair auditable datasets from the movement data. The underlying idea is to first associate a fraction of objects to hotspot(s) of unfairness (recall from Section \ref{sec: preliminaries} that these are the unfair activity-space patterns), and then manipulate these objects' labels such that they are treated differently.
We argue that unfairness can be injected considering several parameters, and in our work we consider the following ones: 
\begin{enumerate} 
\item the \textit{shape} of the regions composing the hotspots; 
\item the \textit{number} of \textit{objects associated} with a hotspot;
\item the \textit{number of hotspots} in a dataset; 
\item the unfairness \textit{magnitude}; 
\item the \textit{number} of \textit{separate regions} composing a hotspot; 
\item how objects are \textit{associated} with the hotspots. 
\end{enumerate}

Parameters 1 through 4 are inspired by protocols defined in the spatial scan statistics literature \citep{kulldorff1997spatial, kulldorff2009scan, jung2010spatial}, while the last two are due to the need to add a activity-space pattern as a possible cause of unfairness. That is, objects can be associated with multiple locations, rather than a single one as previously assumed, and a predictive model can treat objects associated with a certain set of distinct geographical regions differently.
Next, we present the injection protocol we used in this work.

\begin{figure}[t]
\centering
\includegraphics[width=0.457\linewidth]{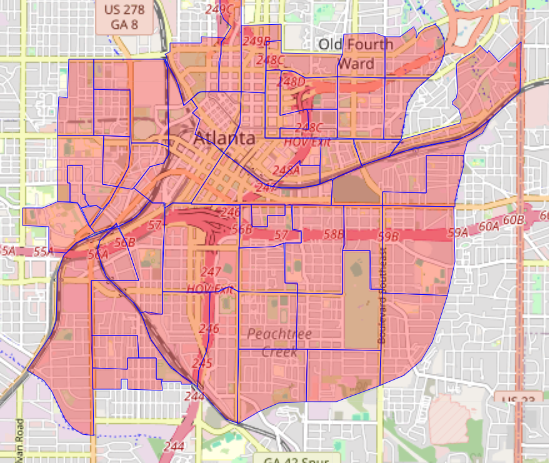}
\includegraphics[width=0.415\linewidth]{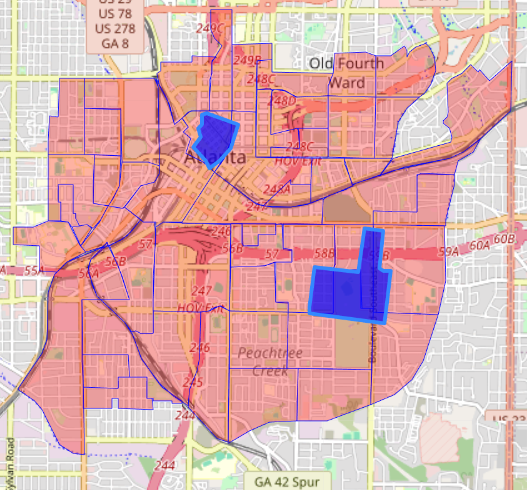}
\caption{\textit{Left plot}: the subset of 51 US Census block groups intersecting the movement data, shown as red-filled with blue edges. \textit{Right plot}: hotspot made of two separate regions (blue filled polygons) constructed from the original block groups.}
\Description{Two maps of the study area. The left panel shows the 51 U.S. Census block groups containing the simulated movement data, represented as semi-transparent red polygons with blue boundaries. The right panel shows an example hotspot constructed from two geographically separate regions, represented by two solid blue polygons.}
\label{fig: hotspot example}
\end{figure}

The first parameter is the shape of the regions composing the hotspots. Shapes should be sufficiently varied and differ from those of the cells of the zonings, to not artificially favor an assessment approach. For this reason, we consider the geometries of a subset of the 2025 US Census block groups for the state of Georgia \citep{uscensus_georgia_2025_blocks}, more precisely the 51 block groups that cover our simulation area (Figure \ref{fig: hotspot example}, left plot). 

More precisely, to construct a hotspot of $n$ distinct regions (which is one of the considered parameters), $n$ distinct block groups are first randomly sampled from the above-mentioned 51, and then undergo the following geometric transformations. Each block group is rotated by a random angle about the centroid of its bounding box, then translated horizontally and vertically by a random offset, subject to the constraint that the transformed polygon still intersects the original's bounding box.
The rotation and translation operations decorrelate the hotspot geometry from the original block group geometry, while the intersection check ensures that the transformed shapes remain geographically localized.

The final transformation we need to apply to the original block groups controls another key parameter, namely the \textit{number of objects associated} with the hotspot. 
This is an important parameter since even when the magnitude of unfairness is kept fixed, a hotspot involving more objects provides a stronger statistical signal, potentially making the unfairness easier to detect but also increasingly harder to delineate its boundaries.

To achieve this, we consider the areas of the $n$ rotated and translated polygons, and apply the same buffer operation to all of them, shrinking or enlarging their areas until the number of associated objects matches a given target value.
Figure \ref{fig: hotspot example}, right plot, provides an example of a hotspot made of two regions derived from the original block groups.
Note that when an exact match is not possible for geometric reasons, we keep the hotspot if the number of objects is sufficiently close to the target. In our evaluation, we allow a maximum absolute difference of 10 objects w.r.t. the target in order to control the variance of this parameter across the generated datasets; otherwise, we discard the hotspot.

Another parameter requires specifying \textit{how objects} become \textit{associated} with a \textit{hotspot}. In our work, we use the centroids of the objects' stop segments computed from their trajectories. More precisely, an object becomes associated with the region of a hotspot if it has at least a given number of stop centroids within that region. An object is therefore associated with the hotspot if it has that number of stop centroids within each hotspot's region. The smallest possible value for this parameter is 1, which requires the object to have stopped at least once in every hotspot region. Larger values impose a stronger association since they require repeated stops in each region. This can make detection and localization easier, but, at the same time, the boundaries of the hotspots become increasingly  more complex, which can negatively impact precise spatial localization.

Another parameter is the \textit{number of hotspots} into which unfairness is injected. To control this parameter, we simply repeat the generation of a single hotspot the desired number of times. 

Finally, the last parameter is the \textit{magnitude} of the injected unfairness, which is controlled by increasing the difference between the distribution of the outcomes of the objects associated with the hotspots and the distribution of the outcomes of the other objects -- for binary outcomes, this is equivalent to increasing the difference between their positive-label rates, since a Bernoulli distribution is fully determined by its success probability.
For example, suppose that the positive label corresponds to a beneficial decision, such as a loan approval. If the classifier assigns this positive label to 60\% of the general population but only to 40\% of the individuals associated with a specific activity-space pattern (hotspot), then that group receives the beneficial outcome 20\% less often, which is precisely the unfairness magnitude. In a group-fairness statistical-parity sense, this is the kind of systematic disparity we want to inject and then detect. If the positive label represented an adverse outcome, the interpretation would be reversed, but the magnitude would still be the difference between the two rates.

We now describe how we generate the unfair auditable datasets used in the experiments. Before going ahead, we clarify that we do not treat the shape parameter as an explicit experimental parameter; instead, each hotspot is generated from block groups that are selected and transformed randomly, as explained before, so variability in this parameter is present within every generated set of datasets. We therefore consider explicitly the other five parameters reported in Table \ref{tab: unfairness injection parameters}, which also lists their corresponding ranges and default values.
We define a \textit{configuration} of unfair auditable datasets as a set of 1,000 datasets independently generated under the same combination of values for these five parameters. Even within a configuration, datasets still differ because the random choices involved in hotspot construction and predicted values generation vary from one dataset to another.

\begin{table}[t]
   \caption{Parameters considered when generating unfair auditable datasets. \label{tab: unfairness injection parameters}}
   
  \centering
  \footnotesize
  \begin{tabular}{lcc}
  & \textbf{Range of values} & \textbf{Default value}\\
  \toprule
    \textbf{Regions per hotspot} & $\{1, 2, 3\}$ & \textbf{2}\\
    \textbf{Stops per region} & $\{1,2,3,4,5\}$ & \textbf{1}\\
    \textbf{Objects per hotspot} & $[200,\ldots,600]$, step $100$ & \textbf{400}\\
    \textbf{Hotspots} & \{1,2,3\} & \textbf{1}\\
    \textbf{Unfairness magnitude} & $[0.2,\ldots,0.6]$, step $0.1$ & \textbf{0.4}\\
    \bottomrule
  \end{tabular}
\end{table}

Overall, we consider 21 configurations, obtained by varying one parameter over its range while keeping the other parameters fixed at their default values.
We note that we pick \textit{one} stop per region and \textit{one} hotspot per dataset as defaults for those parameters in order to consider the minimal nontrivial configuration for unfairness based on activity-space patterns. Using larger values and the resulting increased unfairness complexity are considered when studying those two parameters specifically.

To summarize the experimental loop before turning to the metrics: each auditable dataset is generated by keeping the synthetic movement data fixed, constructing one or more hotspots with known geometry, identifying the objects associated with them, and altering only their predicted values so that their positive-outcome rate differs from that of the remaining objects by the chosen magnitude. Each dataset, therefore, provides a ground truth specifying which are the regions of the hotspots as well as the objects truly treated unfairly. A single configuration consists of 1,000 independently generated datasets sharing the same parameter values, on which we run our assessment approach and record (i) whether it detects unfairness and (ii) which objects it identifies as treated unfairly.

\subsection{Evaluation metrics}
\label{sec: evaluation methodology}

The assessment approach proposed in this work depends on a family of spatial scan statistics mentioned in Section \ref{sec: hypotesis test}. When proposing a new spatial scan statistic, or an approach that uses existing ones (as in our case), one has to evaluate its detection and localization performance.

Detection performance is measured by applying an assessment approach to a group of datasets for which unfairness based on activity-space patterns is present, and then computing the \textit{fraction} of datasets in which the approach successfully \textit{detects} it. This is also known as estimating the \textit{statistical power} of the approach. This requires a \textit{large number} of datasets, and justifies the need to create several distinct groups of synthetic auditable datasets, where unfairness is \textit{injected} considering several parameters to evaluate the detection performance of an assessment approach under different conditions. 

Estimating the statistical power, however, is not enough because it does not quantify how well an approach retrieves the set of objects treated unfairly or how well it \textit{spatially localizes} unfairness. This issue is closely related to the fact that the true shapes of the hotspots of unfairness are typically unknown, as already mentioned when discussing the MAUP in Section \ref{sec: preliminaries}. Moreover, the fact that a hotspot can be composed of multiple regions further complicates the problem.
We therefore borrow two metrics from spatial scan statistics literature, which are used as proxies to evaluate localization performance: \textit{sensitivity} and \textit{Positive Predictive Value (PPV)} \citep{read2011measuring, kulldorff2009scan, jung2010spatial}. 

Formally, consider a synthetic auditable dataset in which unfairness is present, and we also know the sets of moving objects treated differently; 
consequently, let $U$ denote the set of moving objects truly treated differently, and let $\hat{U}$ denote the set of objects an assessment approach considers treated differently. Note that both $U$ and $\hat U$ may contain moving objects from \emph{different} hotspots -- in this case, we assume that $U$ and $\hat U$ simply represent the set union of the objects associated with the hotspots and the candidates deemed as extreme from the hypothesis test in Section \ref{sec: hypotesis test}, respectively. Then, sensitivity measures the fraction of objects truly affected that were correctly detected by an assessment approach, while PPV measures the fraction of objects detected by an assessment approach that truly belong to $U$:
$$\text{\textit{Sensitivity}} = \frac{|U \cap \hat{U}|}{|U|} \quad \quad \quad \text{\textit{PPV}} = \frac{|U \cap \hat{U}|}{|\widehat{U}|}$$
These can be viewed as reinterpretations of the classic recall and precision measures, respectively, because they are applied to the objects associated with the hotspots rather than to the hotspots themselves.

In our experimental evaluation, we compute both measures as averages across a given configuration of unfair auditable datasets, but \textit{limited} to the datasets in the configuration in which the approach \textit{successfully} detected unfairness. We follow this convention from existing literature, e.g. \citep{jung2010spatial, jung2015nonparametric}, to separate the analysis of sensitivity and PPV from that of power. 
To help build intuition for these metrics, we refer the reader to the toy example provided in Appendix \ref{app: supplemental example sensitivity ppv}.

\subsection{Configuration of our assessment approach}
\label{sec: exp eval parameters}

We apply the \textit{trajectory segmentation} step (Section \ref{sec: traj segmentation}) on the compressed trajectories to partition them into stop and move segments. 
Since the simulator generates routine urban mobility, we argue that this step should identify meaningful stays at recurrent locations such as home, work, shopping, or dining places, rather than short pauses during movement. Accordingly, and in the spirit of stay-point formulation from \citet{hariharan2004project}, we set the maximum distance from the stay anchor point to 50 meters and the minimum stay duration to 10 minutes.
Table \ref{tab: movement data charachteristics} summarizes the characteristics of the synthetic movement data and the stop segments.
\begin{table}[h]
   \caption{Movement data and stop segments characteristics. \label{tab: movement data charachteristics}}
   
  \centering
  \footnotesize
  \begin{tabular}{lr}
  \toprule
    \textbf{Objects (trajectories)} & 100,000\\
    \textbf{Sampling rate} & 2 minutes\\
    \textbf{Temporal interval} & 10 days\\
    \textbf{Number of trajectory samples} &  720,199,539\\
    \textbf{Area of the trajectories' bounding box} & $(4.69 \times 4.83) km^2$\\
    \midrule
    \textbf{Daily median number of stops per object} & 4.6 $\pm$ 1.43 \\
    \textbf{Stop median duration} & 122 $\pm$ 472.49 minutes\\
    \bottomrule
  \end{tabular}
\end{table}

Concerning the geographic zonings (Section \ref{sec: space partitioning}), we use uniform grids with square cells.
The set of resolutions was chosen to cover the range of geographic scales that are meaningful in the study area, following the guidelines in Section \ref{sec: space partitioning}. At the lower end, 50 meters approximates the scale of the smallest relevant geographic units, which we consider to be the city blocks. At the upper end, 1000 meters reflects the guideline that a zoning should not be so coarse that a single cell can contain more than half of the population -- in our setting, this corresponds to more than half of the population having at least a stop segment falling within the same cell. The intermediate resolutions were chosen to progressively cover the range between these two extremes, with finer spacing at smaller scales and coarser spacing at larger ones. This choice reflects the fact that changes in resolution matter more at fine geographic scales, while at coarser scales the goal is to capture broader spatial aggregations. 
Finally, for each resolution $X$, we shift the grid boundaries rightward and upward by an offset proportional to the resolution, i.e., $\delta \in [0, X)$ with step $X/5$, obtaining 5 grids per resolution. In total, we use 40 different grids.

The \textit{Object-to-cells mapping} step (Section \ref{sec: users to cells mapping}) maps each object to its top-$i$ (where $i=8$) cells for each considered grid. We believe that this value is reasonable for the expected number of meaningful places in people’s routines, with some margin for variability. 
Finally, we instantiate the \textit{hypothesis testing} step (Section \ref{sec: hypotesis test}) for the case of assessing binary classifiers, i.e., the step uses the Bernoulli-based spatial scan statistic from \citet{kulldorff1997spatial}. Appendix \ref{app: supplemental hyp test binary class} provides the details of such instantiation. In line with previous spatial scan statistics literature, e.g.   \citep{read2011measuring, kulldorff2009scan, jung2010spatial}, we set the statistical significance level to $\alpha = 0.01$ and the number of Monte Carlo simulations to $n=999$.

\section{Experimental evaluation}
\label{sec:experimental evaluation}

In this Section, we evaluate the statistical power, sensitivity, and PPV that our approach achieves across several configurations of datasets. We analyze the results from two complementary perspectives.

The first is a \textit{parameter-wise} perspective, which focuses on how the parameters of the unfairness injection protocol affect the performance of our approach. In this perspective, we compare configurations that differ in one protocol parameter while keeping the grid resolution fixed. In the plots, this corresponds to comparing the considered metrics at a given grid resolution, and allows us to study how power, sensitivity, and PPV change when varying a specific parameter. Connected to this perspective, in each batch of experiments, we also report the results obtained by pooling the extreme candidates detected across all grids. This serves the purpose of capturing the advantages and disadvantages of combining evidence across all considered grid resolutions.

The second is a \textit{resolution-wise} perspective, which focuses on the contribution of each grid resolution to the same three metrics. In this perspective, we follow the behavior of each configuration across grid resolutions. In the plots, this corresponds to reading each curve from finer to coarser grid resolutions. This allows us to study which resolutions favor unfairness detection, which ones favor the retrieval of objects treated unfairly, which ones favor the spatial localization of unfairness, and the various trade-offs one encounters.

\textbf{How to read the figures in the subsequent sections.} Each of Figures \ref{fig: exp num objs} --\ref{fig: exp num stops} reports the results achieved when focusing a specific injection parameter, and shows them through three plots: statistical power, sensitivity and PPV. In every plot, the horizontal axis is the grid resolution and each curve corresponds to one value of the parameter: the parameter-wise view compares curves at a fixed resolution, while the resolution-wise view follows a single curve across resolutions. The table accompanying each Figure reports the same three metrics when the extreme candidates detected across \emph{all} grids are \emph{pooled}. 
Finally, recall from Section \ref{sec: evaluation methodology} that power is estimated over all 1,000 datasets of a configuration, whereas sensitivity and PPV are averaged only over the datasets in which unfairness was actually detected, so the three metrics must not be interpreted with a common denominator.

Across all five injection parameters from Table \ref{tab: unfairness injection parameters} that are varied during the experiments, one pattern is always present, and we state it here once so that the following subsections can be read as confirmations of, or deviations from, it: pooling the extreme candidates from all grids maximizes power and sensitivity but lowers PPV; coarser resolutions individually favor power and sensitivity, while finer resolutions favor PPV. 
Intuitively, aggregating evidence, either by using coarser cells or pooling across grids, makes unfairness easier to detect and its objects easier to retrieve, but draws in nearby objects that are not part of a hotspot, which hurts precise localization. The truly informative deviation from this pattern is the non-monotonic behavior observed when varying the number of regions per hotspot, discussed where it occurs.

\subsection{Number of objects per hotspot}
\label{sec: exp varying num objects}

\begin{figure}[t]
\centering
\includegraphics[width=0.49\linewidth]{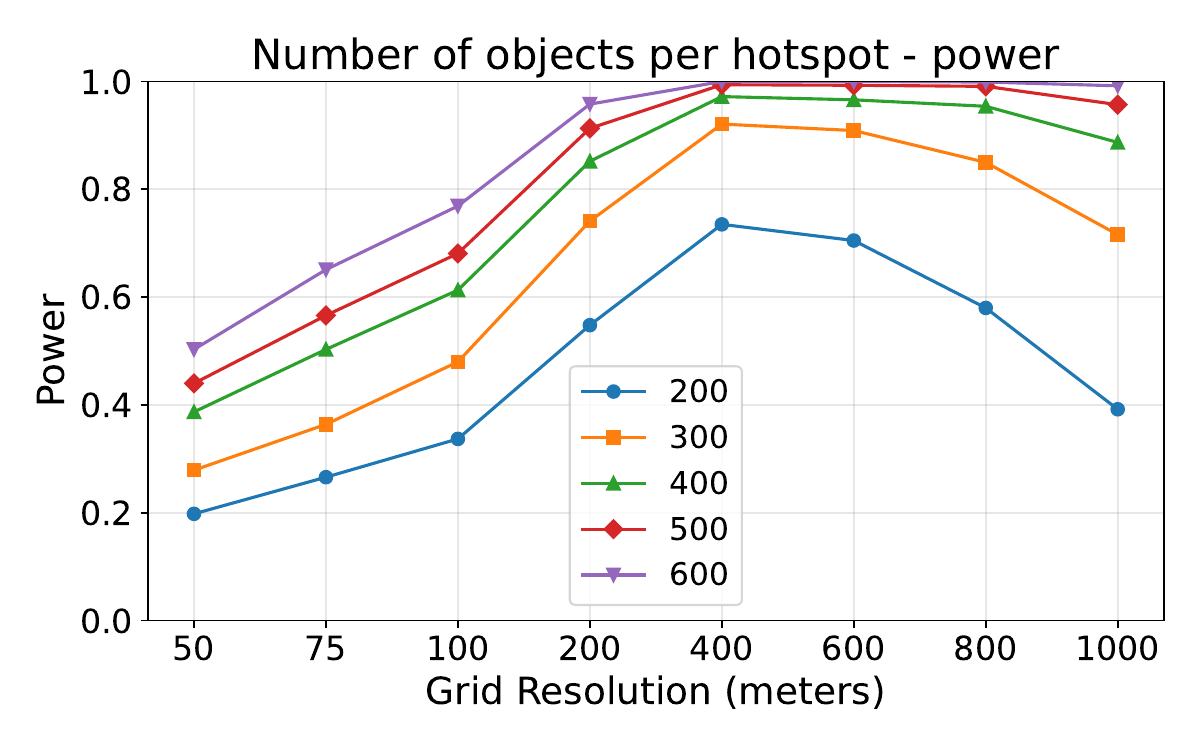}
\includegraphics[width=0.49\linewidth]{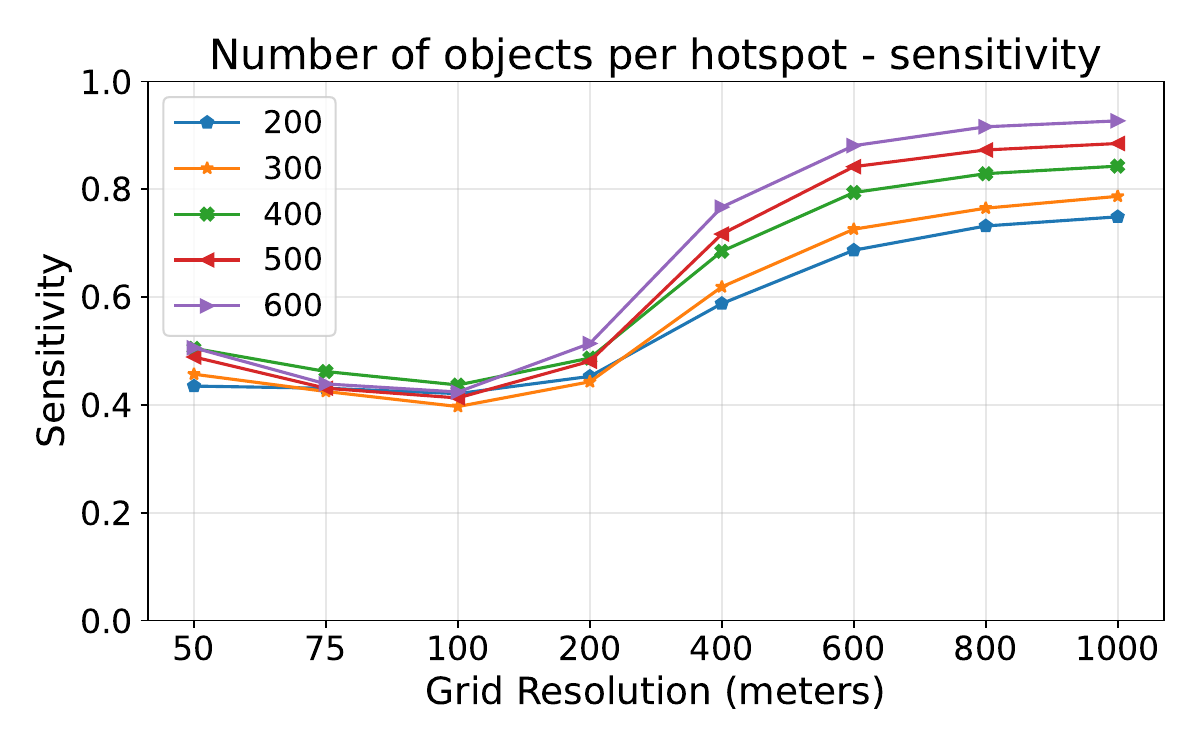}
\includegraphics[width=0.49\linewidth]{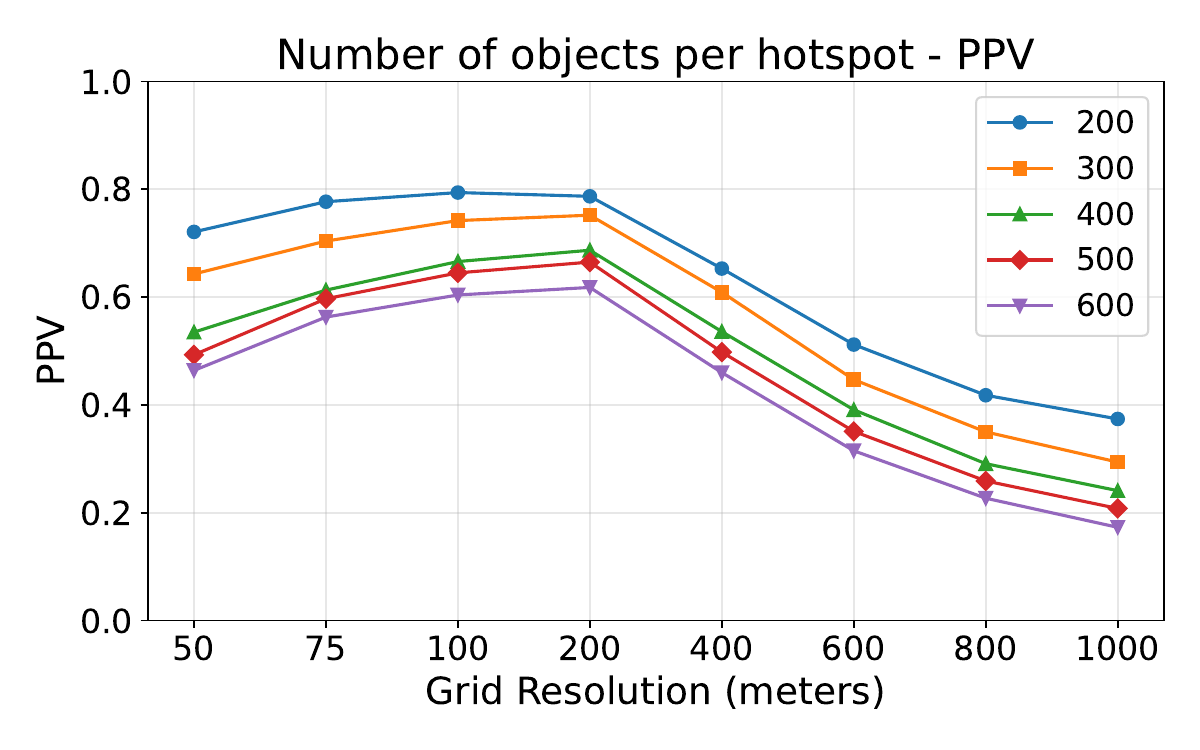}
\caption{Varying the number of objects per hotspot.}
\Description{Three line charts showing power, sensitivity, and positive predictive value, as functions of grid resolution from 50 to 1000 meters, for hotspots associated with 200, 300, 400, 500, or 600 objects. Power generally increases with the number of objects and reaches its highest values at intermediate-to-coarse resolutions. Sensitivity also tends to improve at coarser resolutions and is slightly higher for larger hotspots. PPV shows the opposite behavior: it is highest at finer or intermediate resolutions and decreases as both grid resolution and the number of objects per hotspot increase.}
\label{fig: exp num objs}
\end{figure}

We consider five configurations of 1,000 unfair auditable datasets each, where we vary the number of objects associated with a hotspot across the configurations. All the other parameters are kept at their default values (Table \ref{tab: unfairness injection parameters}). As this parameter grows, we expect the unfairness signal to become easier to detect, but precise localization to become harder because hotspots also grow spatially and develop longer boundaries. Results are shown in Figure \ref{fig: exp num objs} and Table \ref{tab:exp num objects all}.

Let us consider statistical power. Parameter-wise, it increases with the number of objects per hotspot, as expected. Furthermore, Table \ref{tab:exp num objects all} shows that pooling the candidates deemed extreme from all grids exhibits the same trend and maximizes power, thus proving to be a beneficial strategy for unfairness detection. Resolution-wise, the best-performing resolution depends on the number of objects. This suggests that increasing this parameter also tends, on average, to increase hotspot extent, thereby shifting the most suitable resolution. This is an MAUP effect: the best resolution is scenario-dependent, and even within one scenario, different hotspots may be better captured at different resolutions. Finally, observe that the best performing resolutions fall between 400 and 800 meters, which we argue is due to the average extent of the generated hotspots.

\begin{table}[h]
\footnotesize
\centering
\caption{Varying the number of objects per hotspot. Power, sensitivity, and PPV when pooling all the extreme candidates from all grids.}
\label{tab:exp num objects all}
\begin{tabular}{cccc}
\toprule
\textbf{Objects per hotspot} & \textbf{Power} & \textbf{Sensitivity} & \textbf{PPV} \\
\midrule
\textbf{200} & 0.904 & 0.773 & 0.384 \\
\textbf{300} & 0.986 & 0.866 & 0.262 \\
\textbf{400} & 0.998 & 0.926 & 0.193 \\
\textbf{500} & 1.000 & 0.956 & 0.162 \\
\textbf{600} & 1.000 & 0.975 & 0.135 \\
\bottomrule
\end{tabular}
\end{table}

Let us now consider sensitivity. Parameter-wise, we see that it remains relatively stable or slightly increases as the number of objects grows. Pooling the extreme candidates from all grids exhibits the same trend. Furthermore, it maximizes this metric, thus proving to be beneficial for the retrieval of the objects treated unfairly. Resolution-wise, sensitivity tends to improve with coarser resolutions: larger cells are more likely to retrieve a larger fraction of unfairly treated objects.

PPV exhibits opposite trends. Parameter-wise, it decreases as the number of objects per hotspot increases. Moreover, pooling the extreme candidates from all grids proves to be even more detrimental, as it ends up considering even more objects that are not treated unfairly as such. 
A plausible explanation is that, although larger hotspots are easier to detect, their longer boundaries make it more likely that candidates deemed as extreme include nearby objects that are not actually associated with the hotspots, thus reducing spatial localization precision. Resolution-wise, PPV tends to worsen with coarser resolutions, which can be explained by observing that larger cells are more likely to include objects that do not belong to the hotspots.

Overall, the best trade-off among power, sensitivity, and PPV depends on the number of objects per hotspot and grid resolution. This should not be read as evidence that some resolutions are better, but that different resolutions better match hotspots with different characteristics. Moreover, when candidates are progressively added from the finest to the coarsest resolutions, power and sensitivity increase, but PPV does not. In other words, unfairness detection and the retrieval of objects treated unfairly benefit from this strategy, but at the cost of a negative impact on the spatial localization of unfairness.

\subsection{Unfairness magnitude}
\label{sec: exp varying unfairness magnitude}

\begin{figure}[h]
\centering
\includegraphics[width=0.49\linewidth]{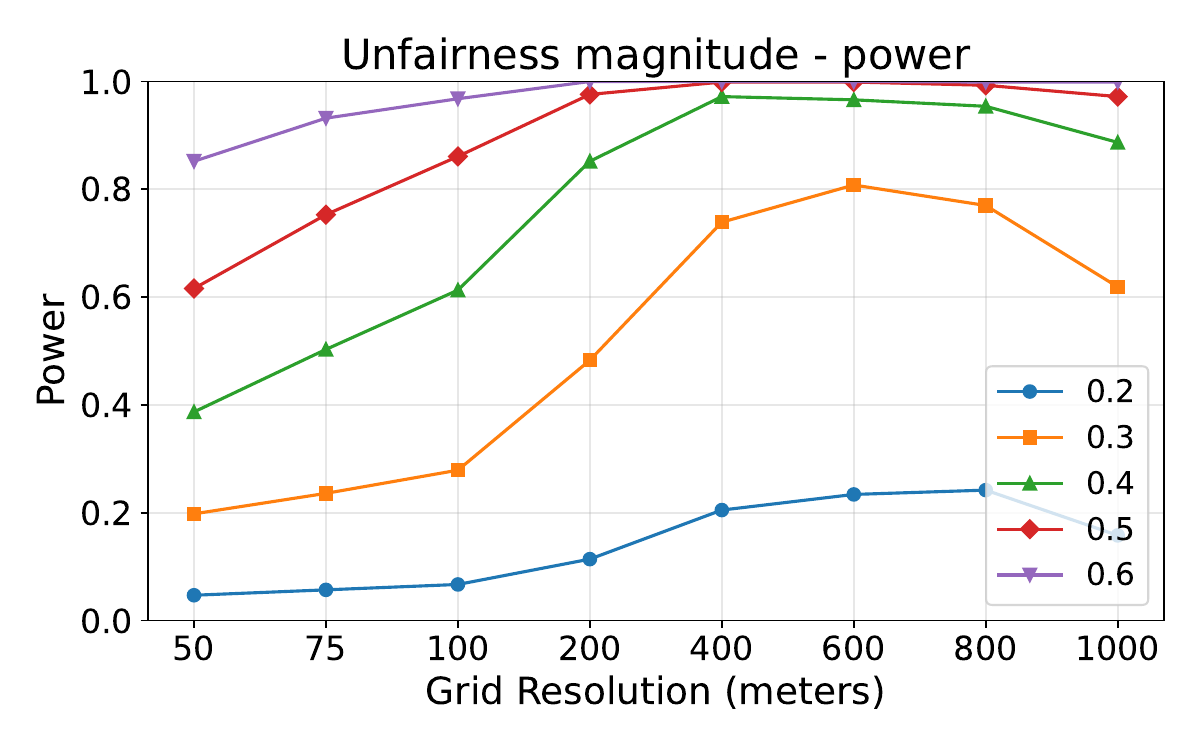}
\includegraphics[width=0.49\linewidth]{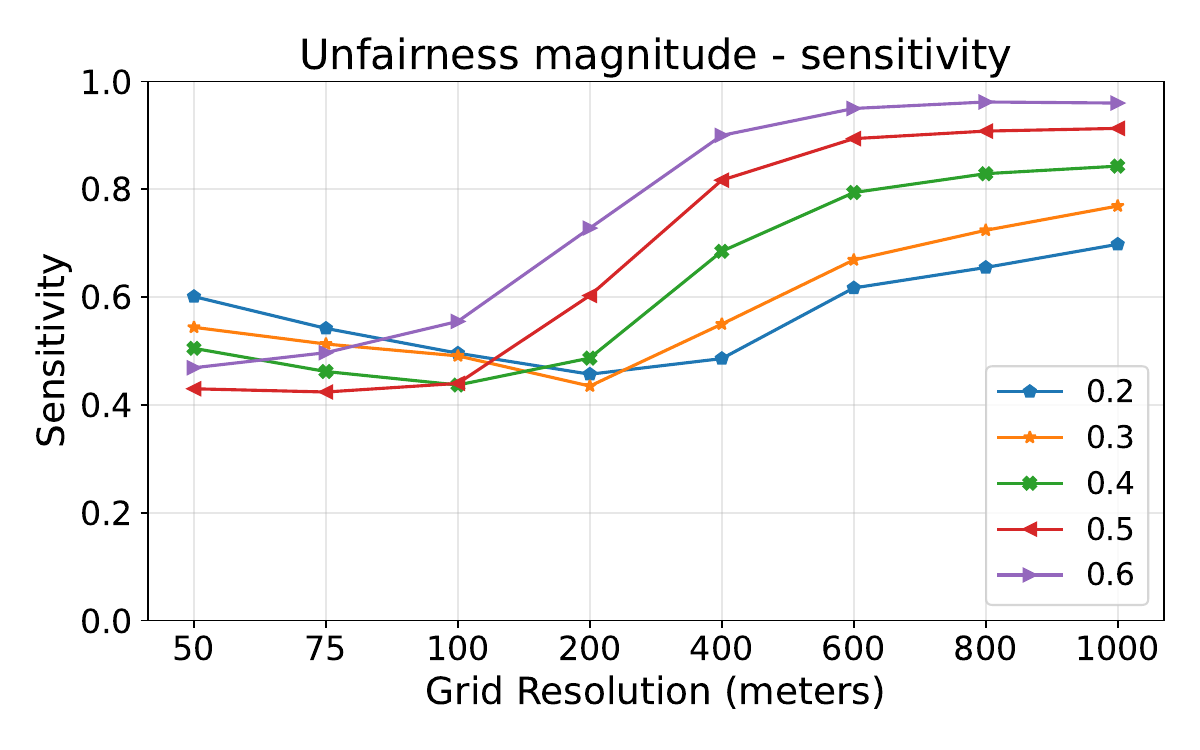}
\includegraphics[width=0.49\linewidth]{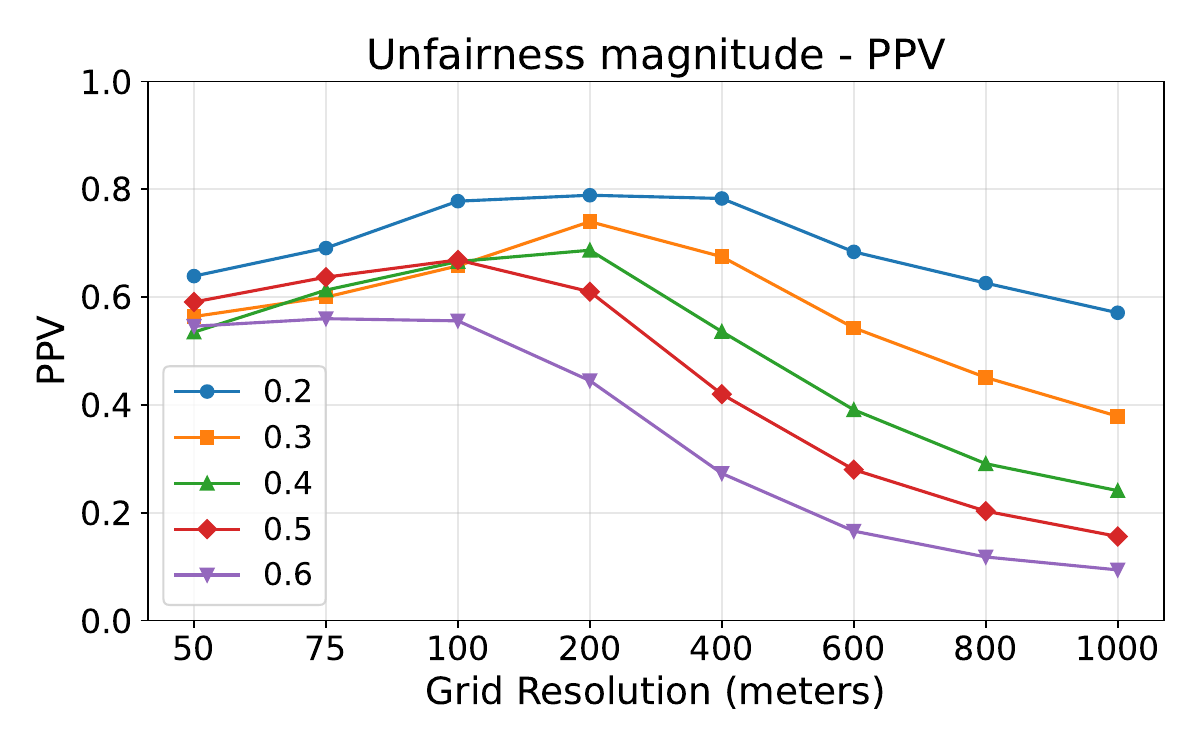}
\caption{Varying the unfairness magnitude.}
\Description{Three line charts showing power, sensitivity, and positive predictive value, as functions of grid resolution from 50 to 1000 meters, for unfairness magnitudes from 0.2 to 0.6. Power increases strongly with unfairness magnitude, with larger magnitudes producing high power even at relatively fine resolutions. Sensitivity generally increases with both unfairness magnitude and grid resolution. PPV instead tends to decrease as unfairness magnitude increases and is generally highest at finer resolutions, especially around 100 to 200 meters.}
\label{fig: exp mag unfair}
\end{figure}

We consider the five configurations of 1,000 unfair auditable datasets each, in which we vary the magnitude of the unfairness injected into the hotspots.
We expect stronger unfairness to be easier to detect, but also to make boundary localization harder, because candidate subsets of cells intersecting hotspot boundaries may increasingly be judged unfair. Results are shown in Figure \ref{fig: exp mag unfair} and Table \ref{tab:exp mag unfair all}.

\begin{table}[h]
\footnotesize
\centering
\caption{Varying the unfairness magnitude. Power, sensitivity, and PPV when pooling all the extreme candidates from all grids.}
\label{tab:exp mag unfair all}
\begin{tabular}{cccc}
\toprule
\textbf{Unfairness magnitude} & \textbf{Power} & \textbf{Sensitivity} & \textbf{PPV} \\
\midrule
\textbf{0.2} & 0.451 & 0.671 & 0.595 \\
\textbf{0.3} & 0.942 & 0.816 & 0.352 \\
\textbf{0.4} & 0.998 & 0.926 & 0.193 \\
\textbf{0.5} & 1.000 & 0.975 & 0.117 \\
\textbf{0.6} & 1.000 & 0.993 & 0.065 \\
\bottomrule
\end{tabular}
\end{table}

Let us consider statistical power. Parameter-wise, it increases with magnitude at every grid resolution, which is expected. The same happens when pooling the extreme candidates from all grids, which also maximizes power thus proving again to be a beneficial strategy for this metric. Resolution-wise, the best performing resolutions concentrate between 400 and 800 meters, which we again argue is due to the average extent of the generated hotspots.

For what concerns sensitivity, parameter-wise it tends to increase as the magnitude increases across most grid resolutions. The same happens when pooling the extreme candidates from all grids, which also maximizes sensitivity thus proving again to be a beneficial strategy for this metric. We argue that as hotspots become more evident, their objects are more likely to be recovered. Resolution-wise, the best performing ones fall above 400 meters, which we argue is due to the average extent of the generated hotspots.

Finally, consider PPV. Parameter-wise, it tends to decrease as unfairness magnitude increases. This decrease is even more evident when extreme candidates from all resolutions are pooled, which again proves that this strategy is detrimental for this metric.
Overall, the results confirm that stronger unfairness not only makes hotspots easier to detect, but also strengthens the signal in candidates intersecting hotspot boundaries; these candidates are therefore more likely to include objects that do not belong to the hotspots, hence lowering PPV, and becomes an especially evident phenomenon with coarser resolutions. Resolution-wise, the best-performing resolutions are the finer ones, i.e., those between 50 and 200 meters. We argue that finer resolutions are likely to include fewer objects that are not treated unfairly, which permits better recognition of the boundaries of the hotspots as unfairness magnitude grows.
Overall, we observe again that coarse resolutions mainly help power and sensitivity, while finer resolutions help PPV.

\subsection{Number of hotspots per dataset}
\label{sec: exp var number hotspots}

\begin{figure}
\centering
\includegraphics[width=0.49\linewidth]{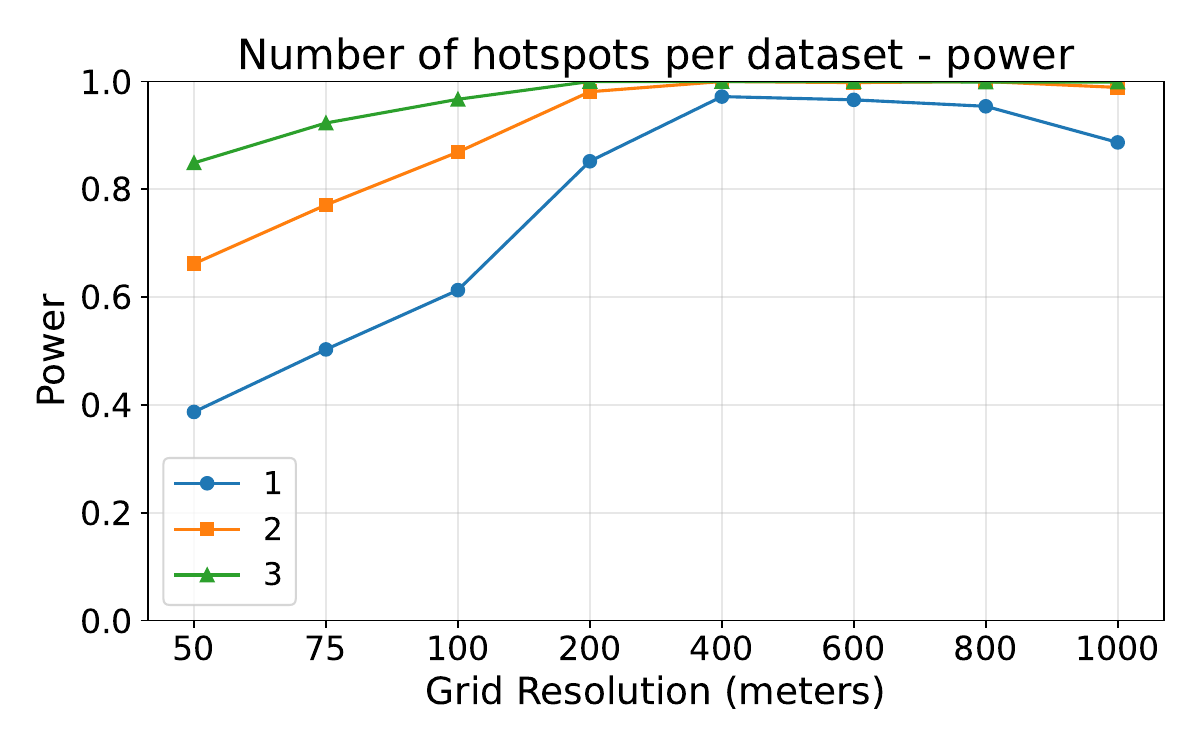}
\includegraphics[width=0.49\linewidth]{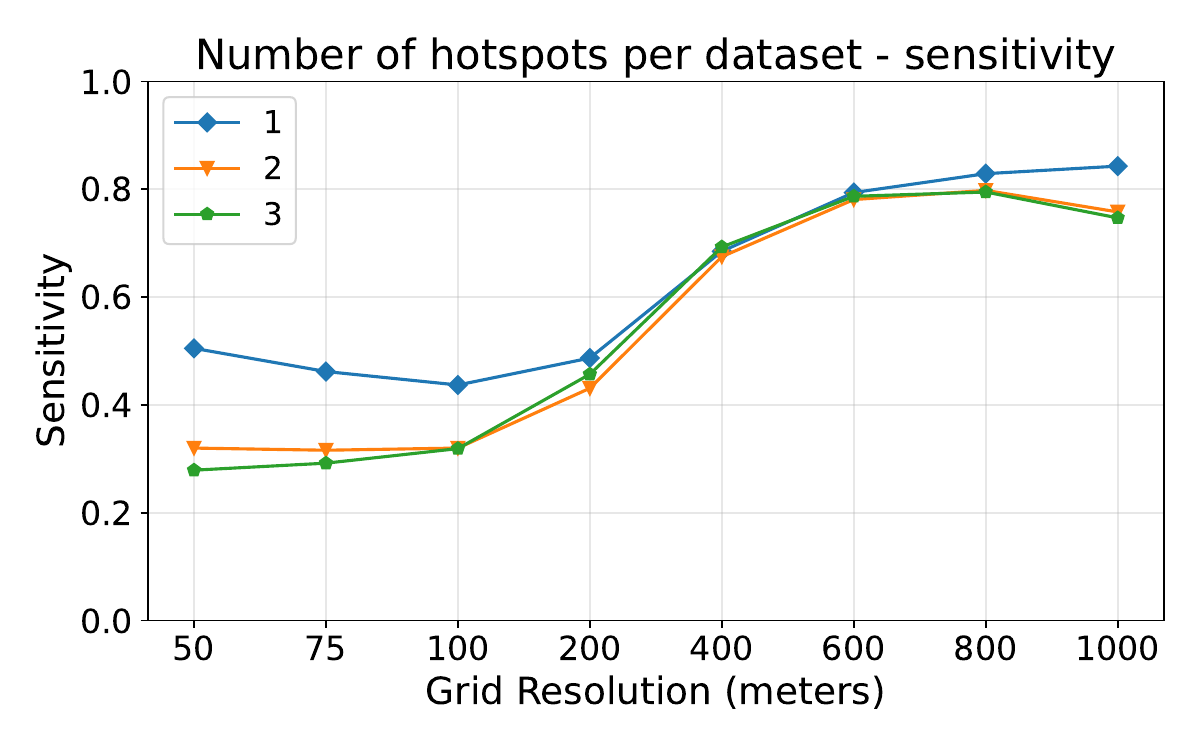}
\includegraphics[width=0.49\linewidth]{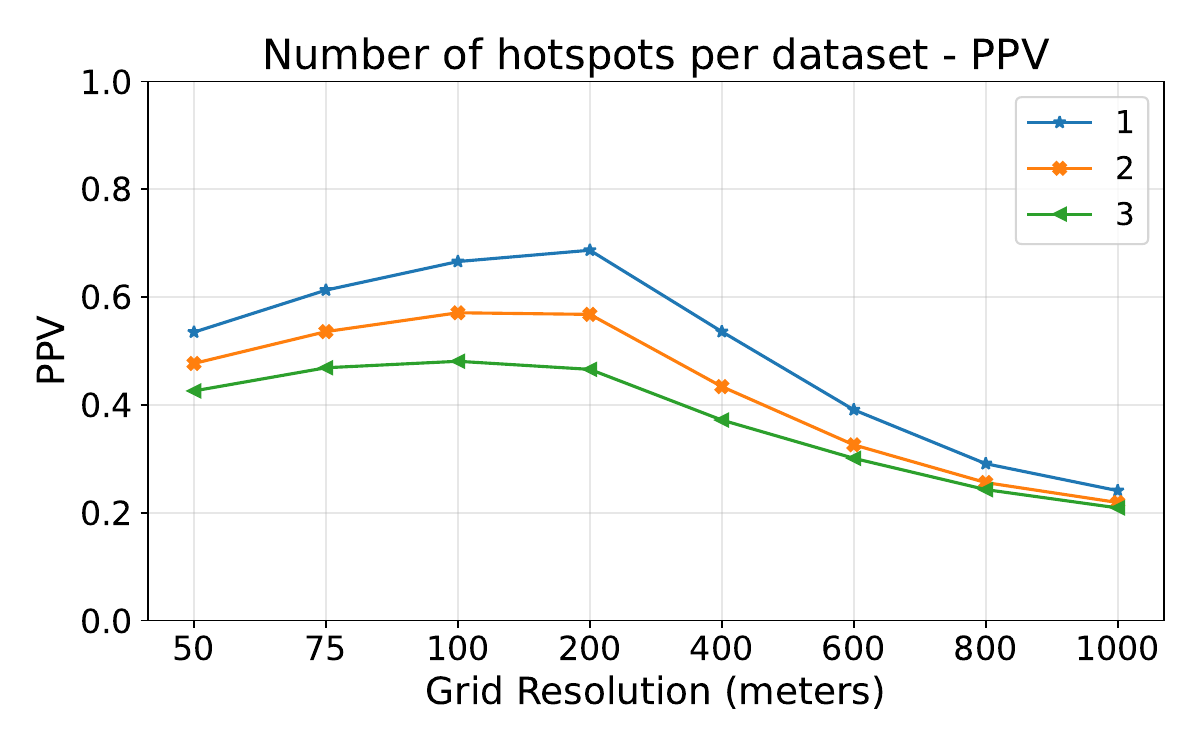}
\caption{Varying the number of hotspots per dataset.}
\Description{Three line charts showing power, sensitivity, and positive predictive value, as functions of grid resolution from 50 to 1000 meters, for datasets containing one, two, or three hotspots. Power increases with the number of hotspots, particularly at fine resolutions, while the curves converge toward high power at coarser resolutions. Sensitivity is lower for multiple hotspots at fine resolutions but becomes similar across configurations at coarser resolutions. PPV decreases as the number of hotspots increases and generally peaks around 100 to 200 meters before declining at coarser resolutions.}
\label{fig: exp num hotspots}
\end{figure}

We now consider the three configurations of 1,000 unfair auditable datasets each, in which we vary the number of hotspots per dataset. 
We expect detection to become easier but localization to become harder, since the approach must deal with hotspots with increasingly more complex boundaries. Results are shown in Figure \ref{fig: exp num hotspots} and Table \ref{tab:exp num hotspots all}.

Let us first consider statistical power. Parameter-wise, power increases with the number of hotspots, as expected. The same trend is observed when pooling the extreme candidates from all grids; in this case, power is also maximized. Resolution-wise, the best-performing resolutions fall between 400 and 800 meters. Furthermore, observe that as we move from finer to coarser resolutions, the gap between the three configurations tends to narrow. We conjecture that finer grid resolutions tend to fragment hotspots more across grid cells. Consequently, adding more hotspots increases the chance that at least one of them, or part of one, aligns well with a grid and yields a statistically significant candidate. With coarser resolutions, instead, even the single-hotspot configuration is often aggregated enough to produce a sufficiently strong unfairness signal. Overall, the results suggest that resolutions that are coarse enough to aggregate the unfairness signal, while not being so coarse as to dilute it, favor unfairness detection.

\begin{table}
\footnotesize
\centering
\caption{Varying the number of hotspots per dataset. Power, sensitivity, and PPV when pooling all the extreme candidates from all grids.}
\label{tab:exp num hotspots all}
\begin{tabular}{cccc}
\toprule
\textbf{Hotspots per dataset} & \textbf{Power} & \textbf{Sensitivity} & \textbf{PPV} \\
\midrule
\textbf{1} & 0.998 & 0.926 & 0.193 \\
\textbf{2} & 1.000 & 0.934 & 0.163 \\
\textbf{3} & 1.000 & 0.934 & 0.155 \\
\bottomrule
\end{tabular}
\end{table}

Sensitivity exhibits different trends. Parameter-wise, for finer grid resolutions below 200 meters, sensitivity tends to decrease moderately as the number of hotspots increases. For coarser resolutions, instead, it remains fairly stable, with only moderate fluctuations. This suggests that as the number of hotspots increases, the retrieval of objects that are treated differently might become more difficult, at least when using certain grid resolutions. When the extreme candidates from all grids are pooled, sensitivity is maximized again. Resolution-wise, the best-performing resolutions are the coarser ones, which suggests again that they represent the better choice when targeting the retrieval of objects that are treated differently.

Finally, consider PPV. Parameter-wise, it decreases as the number of hotspots increases, which is expected. The same is observed when the extreme candidates from all grids are pooled, which again proves to be detrimental to this metric. Resolution-wise, the best-performing resolutions are the finer ones, with the 100 and 200 meter resolutions achieving the highest PPV. The results again suggest that finer resolutions represent the better choice when targeting precise spatial localization of unfairness.

\subsection{Number of regions per hotspot}

\begin{figure}[h]
\centering
\includegraphics[width=0.49\linewidth]{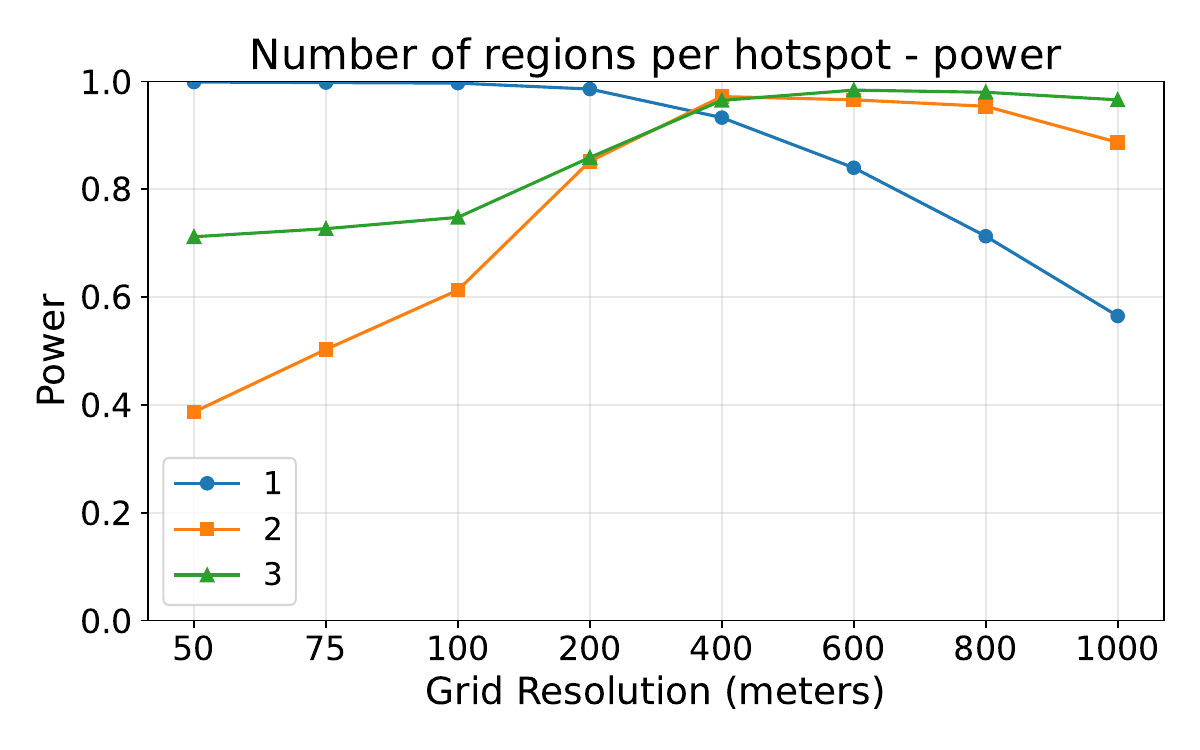}
\includegraphics[width=0.49\linewidth]{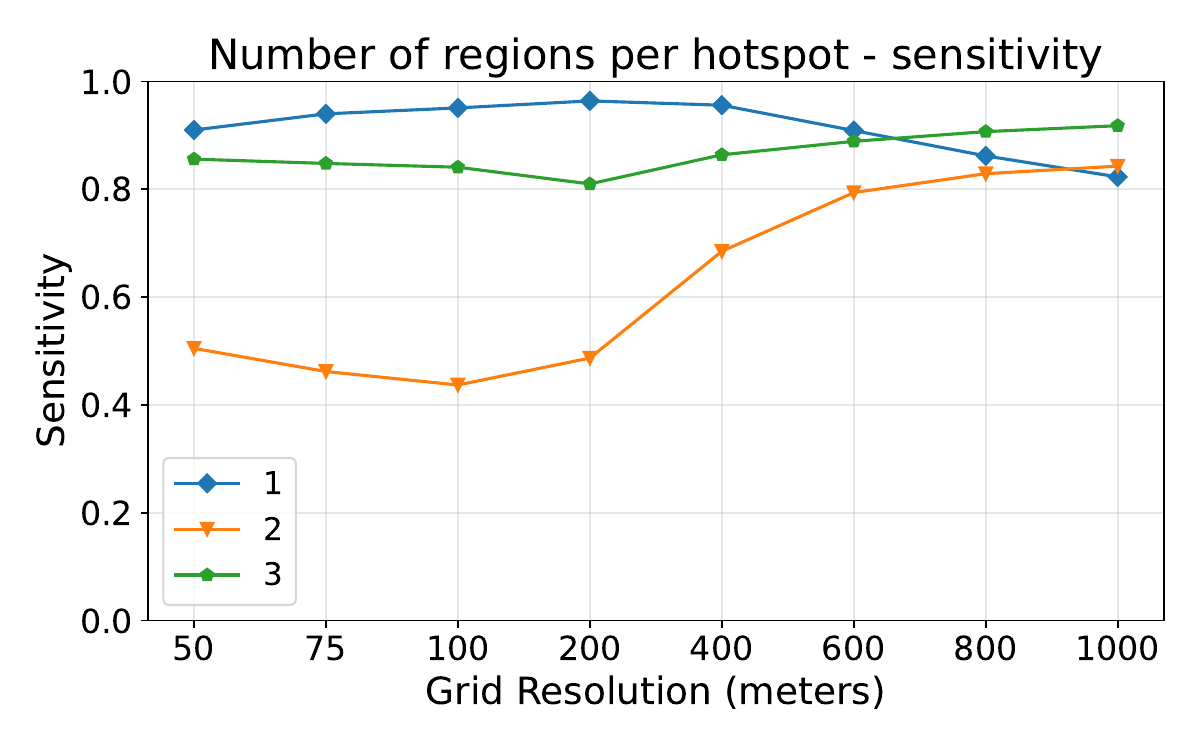}
\includegraphics[width=0.49\linewidth]{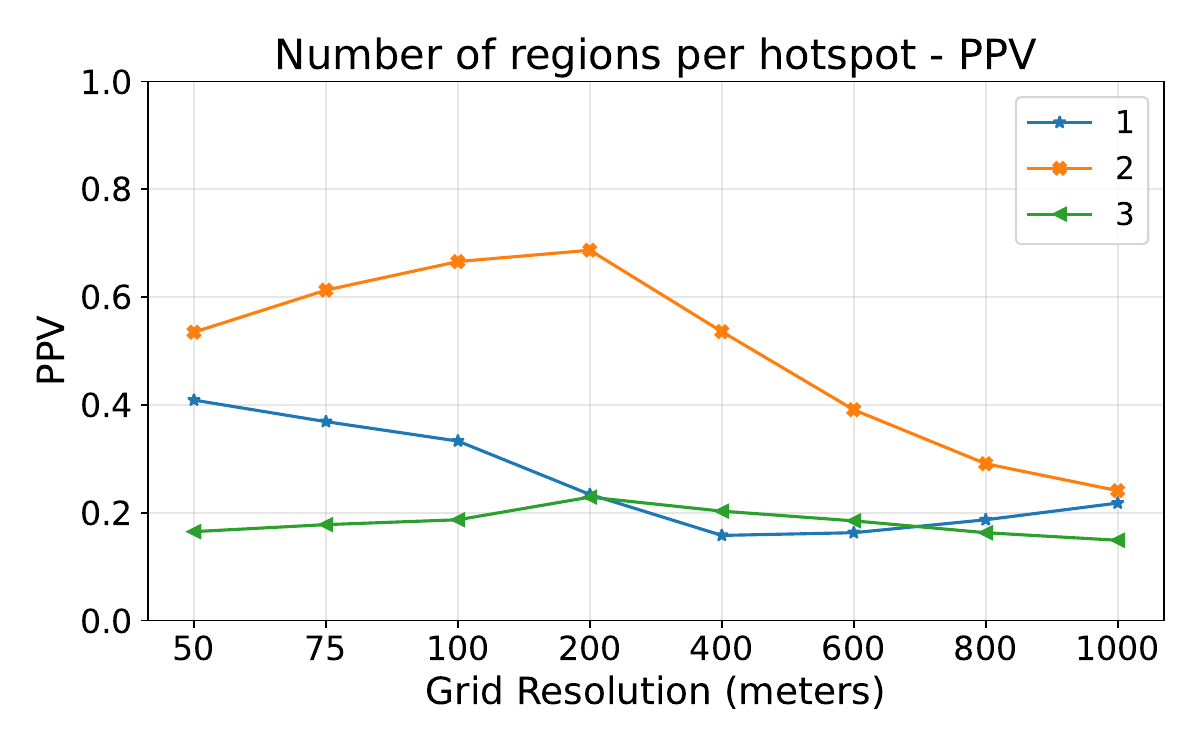}
\caption{Varying the number of regions per hotspot.}
\Description{Three line charts showing power, sensitivity, and positive predictive value, as functions of grid resolution from 50 to 1000 meters, for hotspots composed of one, two, or three separate regions. Power for one-region hotspots is high at fine and intermediate resolutions but decreases at coarse resolutions, whereas power for two- and three-region hotspots generally increases as the grid becomes coarser. Sensitivity remains high for one-region hotspots, increases markedly with resolution for two-region hotspots, and remains relatively high for three-region hotspots. PPV exhibits a non-monotonic dependence on the number of regions: two-region hotspots achieve the highest PPV at fine-to-intermediate resolutions, while one- and three-region hotspots generally achieve lower values. PPV decreases at coarse resolutions in all three cases.}
\label{fig: exp num regions}
\end{figure}

We consider the three configurations of 1,000 unfair auditable datasets each, in which we vary the number of regions composing each hotspot. Increasing the number of regions does not increase the number of associated objects, but it requires those same objects to be associated with more distinct regions; as a result, hotspot regions tend to become larger and their boundaries more complex. Results are shown in Figure \ref{fig: exp num regions} and Table \ref{tab:exp num regions all}.

\begin{table}[h]
\footnotesize
\centering
\caption{Varying the number of regions per hotspot. Power, sensitivity, and PPV when pooling all the extreme candidates from all grids.}
\label{tab:exp num regions all}
\begin{tabular}{cccc}
\toprule
\textbf{Regions per hotspot} & \textbf{Power} & \textbf{Sensitivity} & \textbf{PPV} \\
\midrule
\textbf{1} & 1.000 & 0.990 & 0.105 \\
\textbf{2} & 0.998 & 0.926 & 0.193 \\
\textbf{3} & 0.996 & 0.961 & 0.102 \\
\bottomrule
\end{tabular}
\end{table}

Let us first consider statistical power. Parameter-wise, the effect of increasing the number of regions depends on the grid resolution. At finer resolutions, i.e., below 400 meters, power decreases as the number of regions increases. At coarser resolutions, the opposite trend emerges. Resolution-wise, the one-region case behaves differently from the two- and three-region cases: power decreases as the grid becomes coarser for one-region hotspots, whereas it increases for hotspots composed of two or three regions.
We interpret the results from both perspectives as a scale-matching effect. One-region hotspots are more compact, and are therefore better captured by finer grids. When the grid becomes too coarse, the unfairness signal is diluted by objects that are not truly associated with the hotspot, which reduces power. By contrast, hotspots composed of two or three regions have a larger spatial extent, and in these cases coarser grids are better able to aggregate enough of the objects treated differently to reveal the unfairness signal. Overall, the best-performing resolution tends to become coarser as the number of regions increases. As in the previous experiments, pooling the extreme candidates from all grid resolutions maximizes power.

For sensitivity, parameter-wise the results suggest an interesting non-monotonic effect of the number of regions per hotspot, in part already observed while focusing on statistical power: objects associated with one- or three-region hotspots appear to be generally easier to retrieve than the two-region case. The same is observed when pooling the extreme candidates from all grid resolutions, which again maximizes sensitivity. On the one hand, observe that the unfairness signal of one-region hotspots is \textit{spatially compact}, hence we argue that all grid resolutions, and especially the finer ones, tend to generate candidates that can retrieve most of the objects treated unfairly. On the other hand, the three-region hotspots are spatially more fragmented, which can negatively impact object retrieval, in particular with finer grid resolutions. At the same time, they also provide more regions from which the same objects that are treated unfairly can be retrieved by the candidates deemed extreme, thus favoring object retrieval across all grid resolutions. The two-region case appears to sit in an intermediate situation: it is sufficiently fragmented such that finer grids may split the affected objects across several candidate subsets, but not fragmented enough to provide the same number of retrieval opportunities as the three-region case, leading to the worst results observed for this metric.
Resolution-wise, sensitivity tends to increase with grid resolution for the two-region case, while for the other cases it remains fairly stable, which we argue is due to the reasons explained above. Overall, the results suggest once more that coarser resolutions represent the better choice when retrieving the set of objects that are treated differently.

Let us finally consider PPV. Parameter-wise, the results highlight again a non-monotonic effect of the number of regions per hotspot. We argue that two contrasting phenomena are involved. On the one hand, increasing the number of regions can improve PPV because the unfair activity-space pattern becomes more selective: fewer unaffected objects are likely to be associated with the same set of regions making up a hotspot. On the other hand, as the number of regions increases, hotspots become more spatially fragmented, hence their boundaries become more complex. As a result, candidates deemed extreme may include objects that are not associated with the hotspots, therefore reducing PPV.
Finally, pooling the extreme candidates from all grids proves to be detrimental once more.
Resolution-wise, the best performing resolutions are the finer ones, in line with the previous results.

\subsection{Number of object stops per hotspot region}
\label{sec: exp stops per region}

\begin{figure}[t]
\centering
\includegraphics[width=0.49\linewidth]{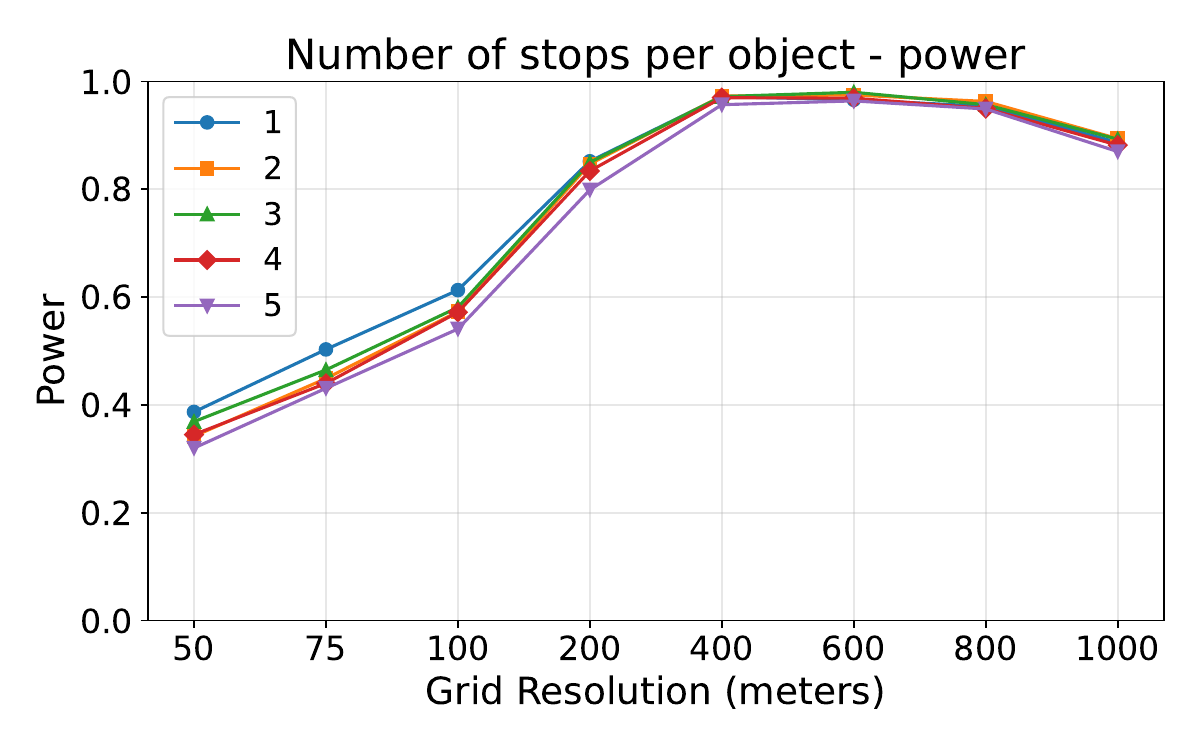}
\includegraphics[width=0.49\linewidth]{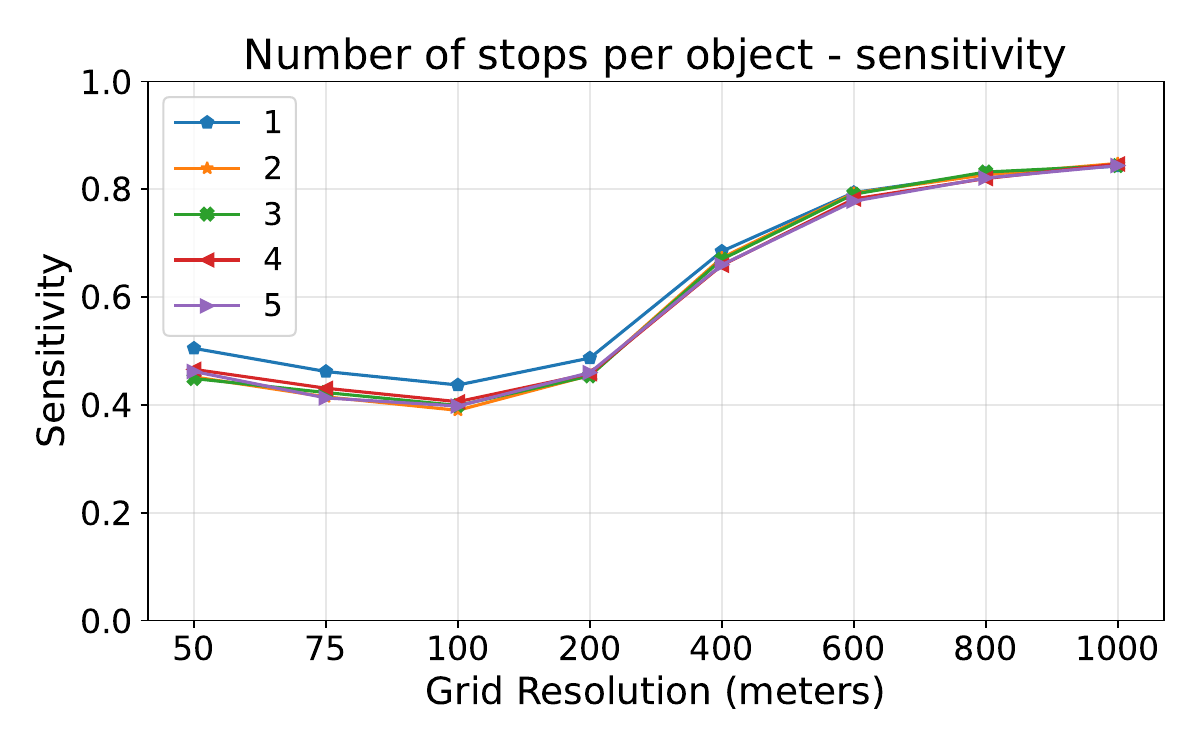}
\includegraphics[width=0.49\linewidth]{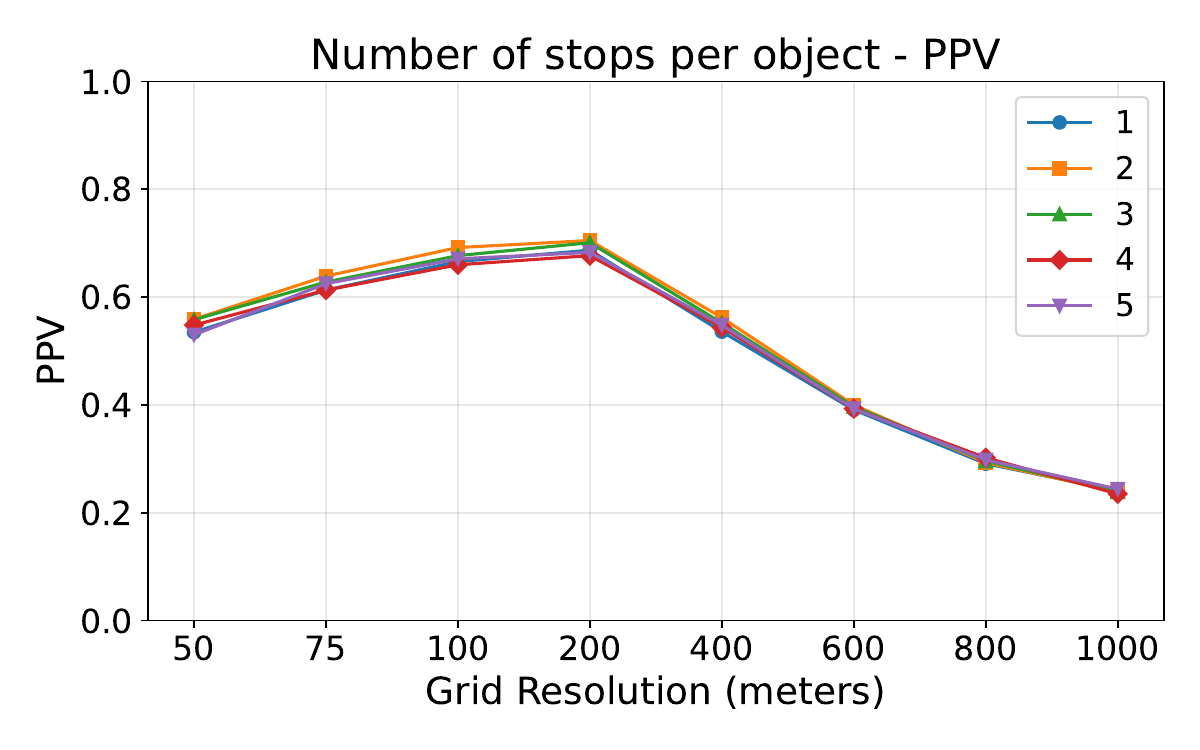}
\caption{Varying the number of stops per hotspot region.}
\Description{Three line charts showing power, sensitivity, and positive predictive value, as functions of grid resolution from 50 to 1000 meters, when requiring from one to five object stops per hotspot region. The curves for the five configurations almost overlap in all three plots, indicating little dependence on the number of required stops. Power increases from fine to intermediate and coarse resolutions, approaching one around 400 to 800 meters. Sensitivity also generally increases toward coarser resolutions after a small decrease at fine resolutions. PPV reaches its highest values around 100 to 200 meters and then decreases as the grid becomes coarser.}
\label{fig: exp num stops}
\end{figure}

We consider the five configurations of 1,000 unfair auditable datasets each, in which we vary the number of stop centroids required for an object to be associated with each hotspot region. 
Increasing this parameter should, on average, enlarge hotspot regions and complicate their boundaries, potentially making both detection and localization harder. On the other hand, it might also facilitate the object-to-cells mapping step of our approach, thus improving performance. Results are shown in Figure \ref{fig: exp num stops} and Table \ref{tab:exp num stops all}.

Let us consider statistical power. Parameter-wise, we observe minimal fluctuations, which suggests that the object-to-cells mapping step from Section \ref{sec: users to cells mapping} is resilient to variations in this parameter. At the same time, this may also depend on properties of the movement data being considered, such as the spatial distribution of the stop centroids. Finally, pooling the extreme candidates from all grid resolutions maximizes power. Resolution-wise, coarser resolutions perform the best, in line with the trends observed in the previous experiments.

\begin{table}
\footnotesize
\centering
\caption{Varying the number of stops per hotspot region. Power, sensitivity, and PPV when pooling all the extreme candidates from all grids.}
\label{tab:exp num stops all}
\begin{tabular}{cccc}
\toprule
\textbf{Object stops per hotspot region} & \textbf{Power} & \textbf{Sensitivity} & \textbf{PPV} \\
\midrule
\textbf{1} & 0.998 & 0.926 & 0.193 \\
\textbf{2} & 0.996 & 0.929 & 0.193 \\
\textbf{3} & 0.999 & 0.927 & 0.196 \\
\textbf{4} & 0.999 & 0.919 & 0.200 \\
\textbf{5} & 0.996 & 0.919 & 0.200 \\
\bottomrule
\end{tabular}
\end{table}

Sensitivity follows similar trends, parameter-wise and resolution-wise. 
Finally, PPV behaves similarly parameter-wise, while resolution-wise the best performing resolutions are the finer ones, which is in line with the results observed in the previous experiments.

\subsection{Discussion}
\label{sec: exp final discussion}

Our evaluation supports three main conclusions. First, unfairness based on activity-space patterns is reliably detected in most configurations, with pooling the extreme candidates from all grid resolutions consistently maximizing statistical power and often bringing it close to or at 1.0. 
Second, retrieving the objects treated unfairly (sensitivity) also generally
benefits from using coarser grids and pooling extreme evidence across grids.
Third, precise spatial localization (PPV) exhibits a trade-off with the previous two objectives: coarser resolutions generally favor power and sensitivity, while finer resolutions generally favor PPV; similarly, pooling maximizes power and sensitivity but lowers PPV, because it draws in nearby objects that are not part of a hotspot. These resolution-dependent trade-offs are consistent with the MAUP in the fairness-assessment setting: there is no single grid resolution or alignment that is uniformly best.

An important practical question at this point is: \textit{what should one do in a real-world assessment?} That is, when one does not know whether there are hotspots, which shapes or spatial scales they have or how best to balance power, sensitivity, and PPV.

To address this question, we outline a practical exploratory strategy based on patterns that recurred throughout the experimental evaluation.
First, we argue that such a strategy should begin by pooling the candidates deemed extreme from all grids. This prioritizes unfairness detection: if no extreme candidate is found at any resolution, then there is no evidence of unfairness based on activity-space patterns. Conversely, if at least one extreme candidate is found, this provides statistical evidence that
unfairness is present, and the analysis must proceed to inspect the objects belonging to the extreme candidates and investigate where the unfairness may be spatially localized.

Next, recall from the experimental evaluation's findings that coarser resolutions mainly support power and sensitivity, while finer ones mainly support precise unfairness spatial localization and thus PPV. Consequently, we suggest beginning with coarse resolutions, which provide a broad indication of where unfair hotspots may be located and facilitate the retrieval of objects treated unfairly. Subsequently, one should turn to finer resolutions to refine those areas and restrict the set of objects deemed as treated unfairly, but only when such resolutions offer a clear and spatially consistent refinement of the same hotspots.
Finally, if hotspots exist at different spatial scales, separate resolution ranges may need to be examined.

We conclude by making the limitations of our study explicit. First, we use uniform square grids; the spatial scan statistics literature has long shown that detection and localization depend on cell shape and on the tessellation used \citep{tango2005flexibly}. This is connected to another limitation observed from the results, i.e., precise spatial localization of unfairness degrades as hotspots grow, since longer or more complex boundaries increasingly involve nearby objects that are not treated unfairly.
Hence, using data-driven or density-aware partitions may improve the considered metrics. 
Second, our approach associates objects with cells through stop segments only, discarding the order of visits and the move segments between stops. We conjecture that incorporating this additional information might enable us to discover other types of unfairness from the mobility of individuals.
Third, in the absence of real auditable datasets our evaluation necessarily uses synthetic ones. Furthermore, our experiments instantiate the approach for the case of assessing a binary classifier. Extending the evaluation to other scenarios (e.g., to regressors or multi-class classifiers) is straightforward in principle but left to future work. 
Fourth, superimposing multiple resolutions and alignments mitigates the MAUP but does not eliminate it, which is precisely why we recommend the practical exploratory strategy outlined in Section \ref{sec: exploratory strategy}. We conjecture that this strategy could be further improved by adopting more principled mechanisms for reporting the candidate subsets deemed extreme, for example, by adapting approaches developed in the spatial scan statistics literature (e.g., using the Gini coefficient, such as in \citep{han2016using}).

Finally, we would like to highlight that we did not report a direct comparison with existing methods assessing predictive models for spatial fairness, as they do not address fairness based on activity-space patterns and therefore are not directly applicable to our setting. The closest point of contact is the reduction presented in Section \ref{sec: reduction to spatial fairness}.

\section{Searching for potential unfairness: a sample case}
\label{sec: exploratory strategy}

We now illustrate how the exploratory strategy outlined in the previous discussion can be used in a real-world assessment. We randomly select one auditable dataset from the experiments but we do not inspect its characteristics, as we aim to verify whether the strategy can correctly guide the assessment, and thus reveal the true hotspots.

\begin{figure*}
\centering
\includegraphics[width=0.32\linewidth]{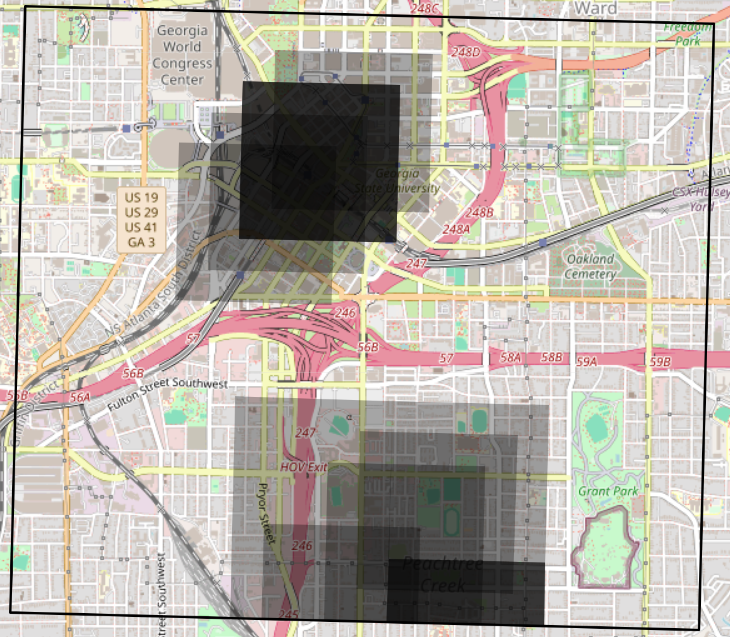}
\includegraphics[width=0.32\linewidth]{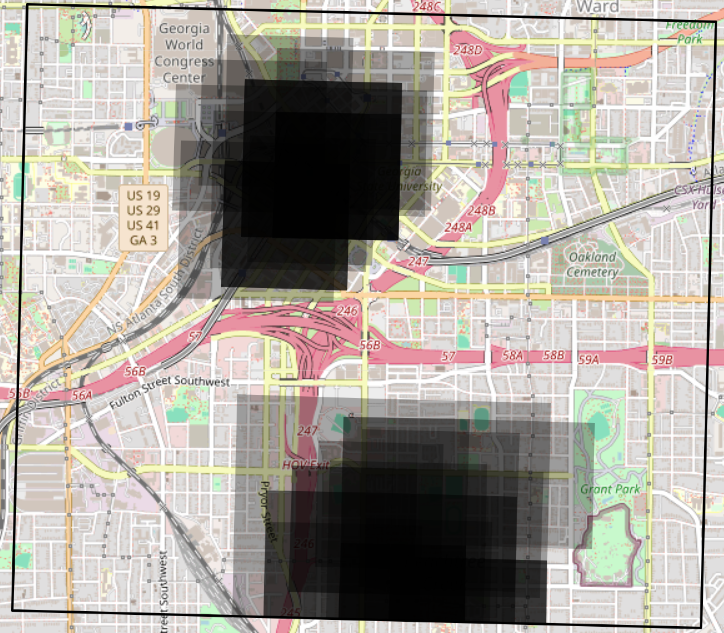}
\includegraphics[width=0.32\linewidth]{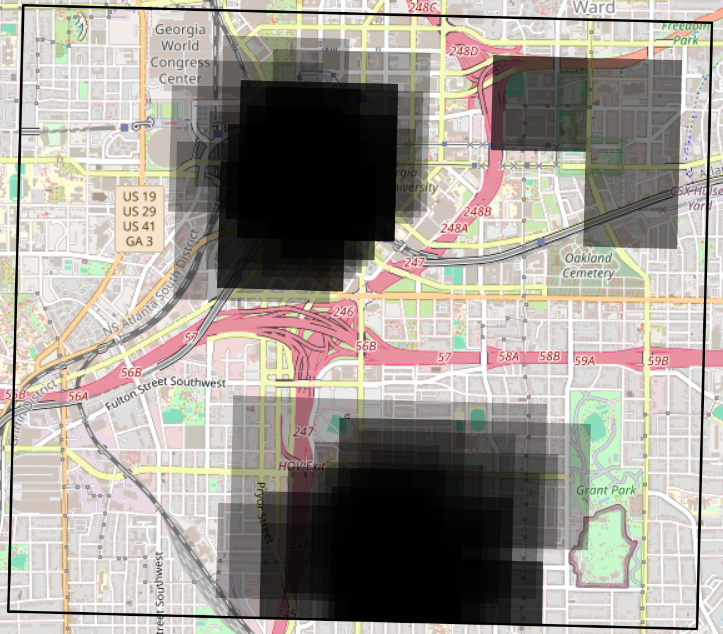}
\includegraphics[width=0.32\linewidth]{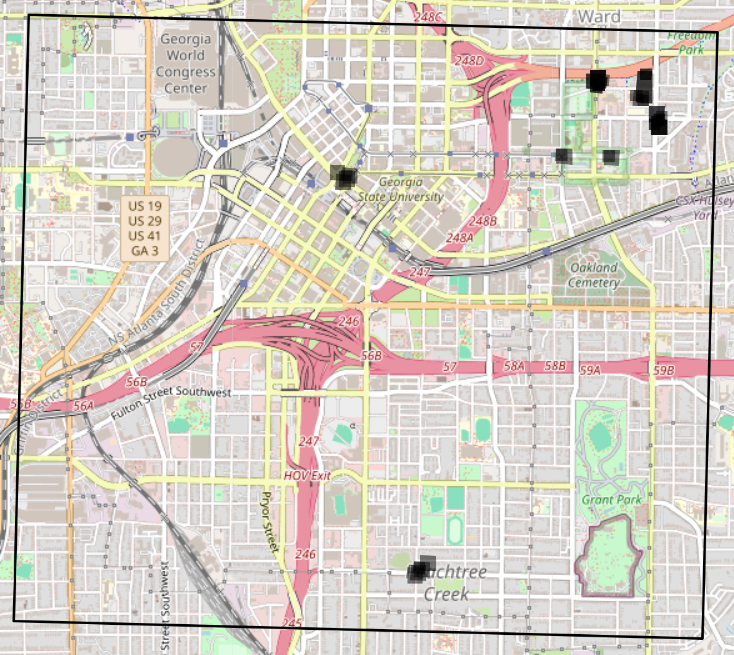}
\includegraphics[width=0.32\linewidth]{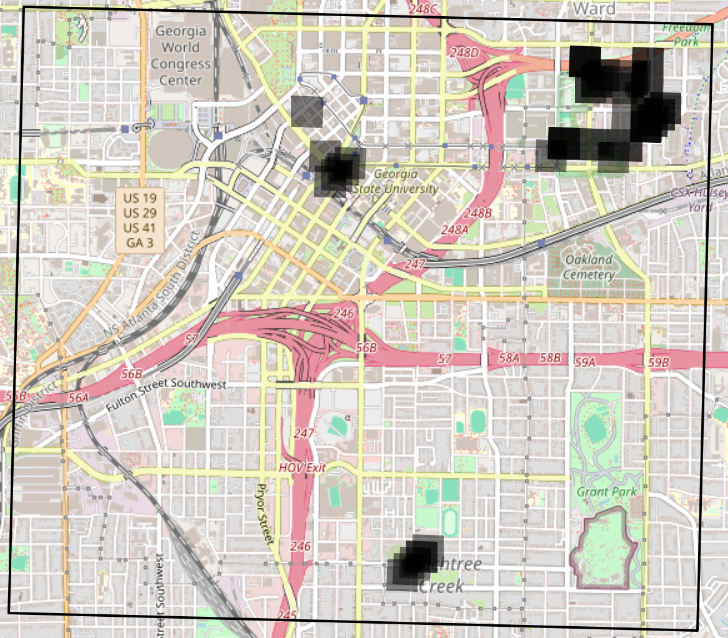}
\includegraphics[width=0.32\linewidth]{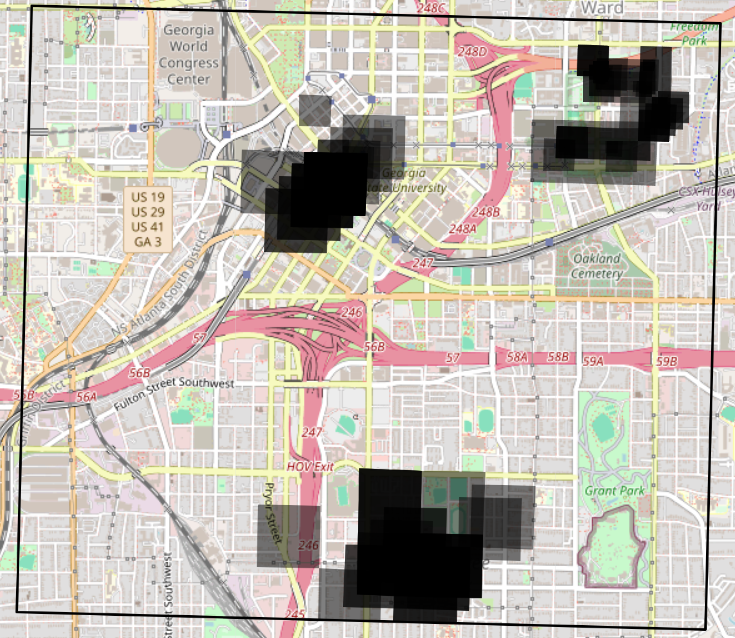}
\caption{The plots in the \textit{top row} show the cells, i.e., the gray-shaded areas, belonging to the extreme candidates detected at coarser grid resolutions. The plots in the \textit{bottom row} show the cells belonging to candidates detected at finer grid resolutions. The darker an area is, the more it is covered by cells from extreme candidates. The black rectangle represents the bounding box containing the movement data.}
\Description{Six maps arranged in two rows of three, showing cells belonging to extreme candidate subsets detected at different grid resolutions. Darker gray areas indicate locations covered by cells from a larger number of extreme candidates, and a black rectangle marks the bounding box of the movement data. The top row progressively pools coarse resolutions: the left plot uses 1000-meter grids and reveals two broad areas; the middle plot adds 800-meter grids and reinforces these areas; the right plot also adds 600-meter grids and reveals a third broad area. The bottom row progressively pools finer resolutions: the left plot combines 50-, 75-, and 100-meter grids and identifies smaller localized areas; the middle plot adds 200-meter grids, causing nearby areas to begin coalescing; and the right plot adds 400-meter grids, producing three larger and more continuous areas.}
\label{fig: cand coarse res}
\end{figure*}

We begin by executing our assessment approach on the dataset, configuring it according to the parameters described in Section \ref{sec: exp eval parameters}.
We first verify whether its final output, \textit{UNFAIR}$_{\mathcal{GZ}} \neq \emptyset$, i.e., contains at least one candidate subset deemed extreme. Observe that this follows the strategy of pooling extreme candidates across all resolutions, which maximizes unfairness detection (i.e., statistical power). Being that the case, we now know that there is statistical evidence that unfairness is present and further investigation is required. 

Next, we consider the extreme candidate subsets detected at coarse resolutions, with the aim of favoring object retrieval (i.e., sensitivity). Using only the 1000-meter grids, two broad areas emerge (Figure \ref{fig: cand coarse res}, top-left). Adding the 800-meter grids makes these areas slightly larger (Figure \ref{fig: cand coarse res}, top-middle), strengthening the conclusion that unfairness is roughly localized there. When the 600-meter grids are also added (Figure \ref{fig: cand coarse res}, top-right), a third broad area appears. This suggests that unfairness may thus be located in these three areas. We now turn to the finer resolutions, with the aim of favoring unfairness spatial localization (i.e., PPV).
Using finer grids, e.g., 50-, 75-, and 100-meter resolution, highlights areas within the larger areas previously identified (Figure \ref{fig: cand coarse res}, bottom-left). Adding the 200-meter grids makes the smaller areas begin to coalesce into larger ones (Figure \ref{fig: cand coarse res}, bottom-middle). Finally, adding the 400-meter grids makes all three areas coalesce further and enlarge (Figure \ref{fig: cand coarse res}, bottom-right).

\begin{figure}
\centering
\includegraphics[width=0.49\linewidth]{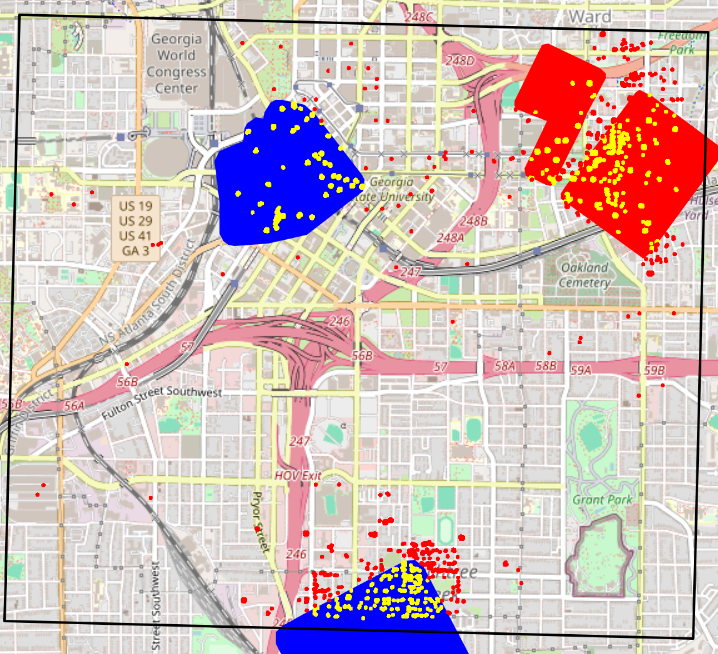}
\includegraphics[width=0.49\linewidth]{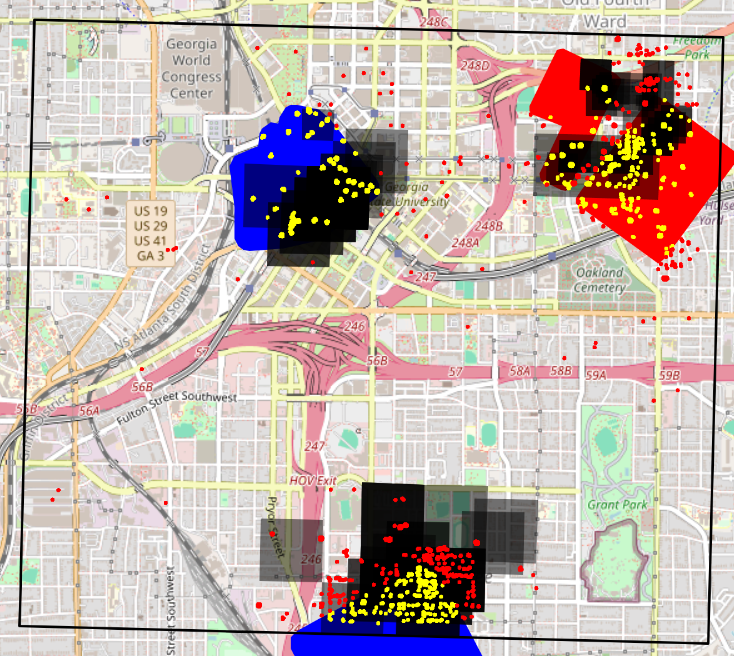}
\caption{The \textit{left plot} shows the two actual hotspots, each composed of two regions, highlighted in blue and red, respectively. The yellow and red points show the stop centroids of the objects associated with the hotspots: yellow points fall within the hotspot regions, while red points fall outside them. The \textit{right plot} superimposes, on the hotspots, the cells belonging to the extreme candidates detected from the 50-, 75-, 100-, 200-, and 400-meter grid resolutions, shown as gray-shaded areas, together with the stop centroids.
\Description{Two maps comparing the true injected hotspots with the areas identified by the assessment approach. In the left plot, two hotspots are shown, one in blue and one in red, and each hotspot consists of two geographically separate polygonal regions. Stop centroids of objects associated with the hotspots are also shown: yellow points are centroids falling inside a hotspot region, while red points are centroids of the same objects falling outside the hotspot regions. The right plot additionally superimposes gray-shaded cells belonging to extreme candidates detected using grid resolutions from 50 to 400 meters. These detected areas substantially overlap the true hotspot regions while also covering some nearby areas containing outside stop centroids of the unfairly treated objects.}
\label{fig: true hotspots}}
\end{figure}

At this point we look at the true hotspots in the auditable dataset, with the intent to verify if our strategy has indeed found out areas in which there is unfairness.
Recall from Section \ref{sec: exp eval unfairness injection} that, in our generation protocol, each hotspot is generated independently and may be composed of multiple distinct regions. Recall also that an object is associated with that hotspot only if it has at least a given number of stop centroids in each of those regions. Hence, in the subsequent plots, each color denotes one specific hotspot while the disconnected polygons with the same color denote the regions composing it.
Figure \ref{fig: true hotspots}, left plot, shows that there are two hotspots, each composed of two distinct regions, highlighted in blue and red respectively. 
The figure also shows the stop centroids of the objects associated with the hotspots: the yellow centroids fall within the hotspot regions, while the red ones fall outside. Figure \ref{fig: true hotspots}, right plot, then superimposes the gray areas, i.e., the cells belonging to the extreme candidates detected from the 50- to the 400-meter grids, over the two hotspots. This plot shows that the three detected areas, in fact, roughly cover the regions of the hotspots.

We finally consider the stop centroids falling outside the hotspot regions, represented by the red dots in Figure \ref{fig: true hotspots}. 
Observe that the detected candidates cover areas outside the true hotspots, which might initially be interpreted as poor spatial localization of unfairness. However, note that part of these areas actually contain many of those outside (red) centroids. In turn, those centroids highlight that unfairly treated objects in reality have broader activity-space patterns than those defined by the hotspot-triggering regions alone. We thus argue that our assessment approach has correctly determined the presence of unfairness in those specific areas. 

Overall, we argue that our practical exploratory strategy provides a reasonable approximation of the areas that warrant deeper investigation.
\section{Conclusions}
\label{sec: conclusions}

In this work, we introduced the problem of assessing predictive models for fairness based on activity-space patterns. We also proposed an approach to address this problem and evaluated it experimentally in the case of assessing binary classifiers. To this end, we first introduced a protocol to generate synthetic auditable datasets, and then evaluated its detection, object retrieval, and localization performance with respect to several parameters. The results suggest that our approach is effective at detecting unfairness based on activity-space patterns and at retrieving the objects treated unfairly, while also providing useful localization performance. 
The results also highlight multi-resolution trade-offs between unfairness detection, object retrieval, and precise localization of unfairness. Accordingly, we suggest an exploratory strategy when dealing with real-world assessments. Finally, we note that our proposed approach can be specialized to address the spatial fairness assessment scenario where  individuals are tied to a single location, thus being generic in nature.

There are multiple directions of future research.
First, it would be useful to move beyond the uniform square grids adopted in our study and previous related works, and investigate alternative tessellations, e.g., data-driven or density-aware ones such as those obtained with DBSCAN \citep{schubert2017dbscan}: these may better reflect the spatial distribution of movement data, thus improving all the considered metrics.
Closely related is the shape of the cells in the tessellations, hence we argue that it would be important to study how different geometries can impact the detection and localization performance when dealing with our problem.

A second direction is to improve the reporting of the candidate subsets of cells deemed extreme, with the goal of improving unfairness spatial localization and benefiting the exploratory strategy. Similar problems have been studied in the spatial scan statistics literature \citep{han2016using}, hence it would be interesting to apply these ideas within our setting.

A third direction concerns how objects are associated with cells. In our work, this is based on stop segments, but alternative criteria, such as those that consider move segments, might be more effective in different scenarios.
Another direction could develop richer and more varied unfairness injection protocols,  investigate movement data with different characteristics, and consider assessing other types of predictive models (e.g., regression, multi-class classification).

\appendix
\begin{appendices}
\newpage

\section{Group fairness targeting statistical parity}
\label{app: supplemental group fairness stat parity}

Let $\mathcal{A} \in \{0,1\}$ represent a binary protected attribute defining a group (e.g., sex, race, income level) and where the values 1 and 0 encode membership or not, respectively, to such a group.
Let also $\hat{Y}$ be the output of a predictive model for some moving object.
Group fairness targeting statistical parity requires 
$$d\big( P(\hat{Y} \ | \ \mathcal{A}=1), \ P(\hat{Y} \ | \ \mathcal{A}=0) \big) \leq \epsilon$$
to hold, where $P(\cdot | \cdot)$ is a conditional probability distribution, $d(\cdot, \cdot)$ is a suitable distance measure between probability distributions (e.g., total variation distance), and $\epsilon$ is a nonnegative threshold that encodes the maximum disparity deemed acceptable, and in practice is an application- and policy-dependent threshold \citep{dwork2018group}.\\

For example, consider a predictive model that is a binary classifier (hence $\hat{Y} \in \{0,1\}$). Then, the above condition reduces to checking if the probability of an unfavorable output (e.g. $\hat{Y}=1$) for an individual belonging to the protected group (i.e., $\mathcal{A}=1$), in other words $P(\hat{Y}=1 \ | \ \mathcal{A}=1)$, is sufficiently close (i.e., within $\epsilon$) to that of an individual of the non-protected group, $P(\hat{Y}=1 \ | \ \mathcal{A}=0)$. 

Finally, we recall from the literature that targeting exact statistical parity, i.e., $\epsilon=0$, is not desirable, as it would sacrifice a model's utility (e.g., accuracy) for total fairness beyond what is reasonable \citep{dwork2018group, friedler2019comparative, Shaham22}. Hence, our approach uses exact statistical parity as the null hypothesis to identify statistically significant disparities in model outcomes. Whether these disparities exceed an application-dependent tolerance $\epsilon$ is a separate question, left to subsequent assessment. For example, in the case of assessing a binary classifier, $\epsilon=0.05$ allows a gap of five percentage points: probabilities of 60\% and 56\% are within tolerance, while 60\% and 52\% exceed it.

\section{The Modifiable Areal Unit Problem: an illustrative example}
\label{app: supplemental MAUP example}

\begin{figure}
\centering
\includegraphics[width=0.45\linewidth]{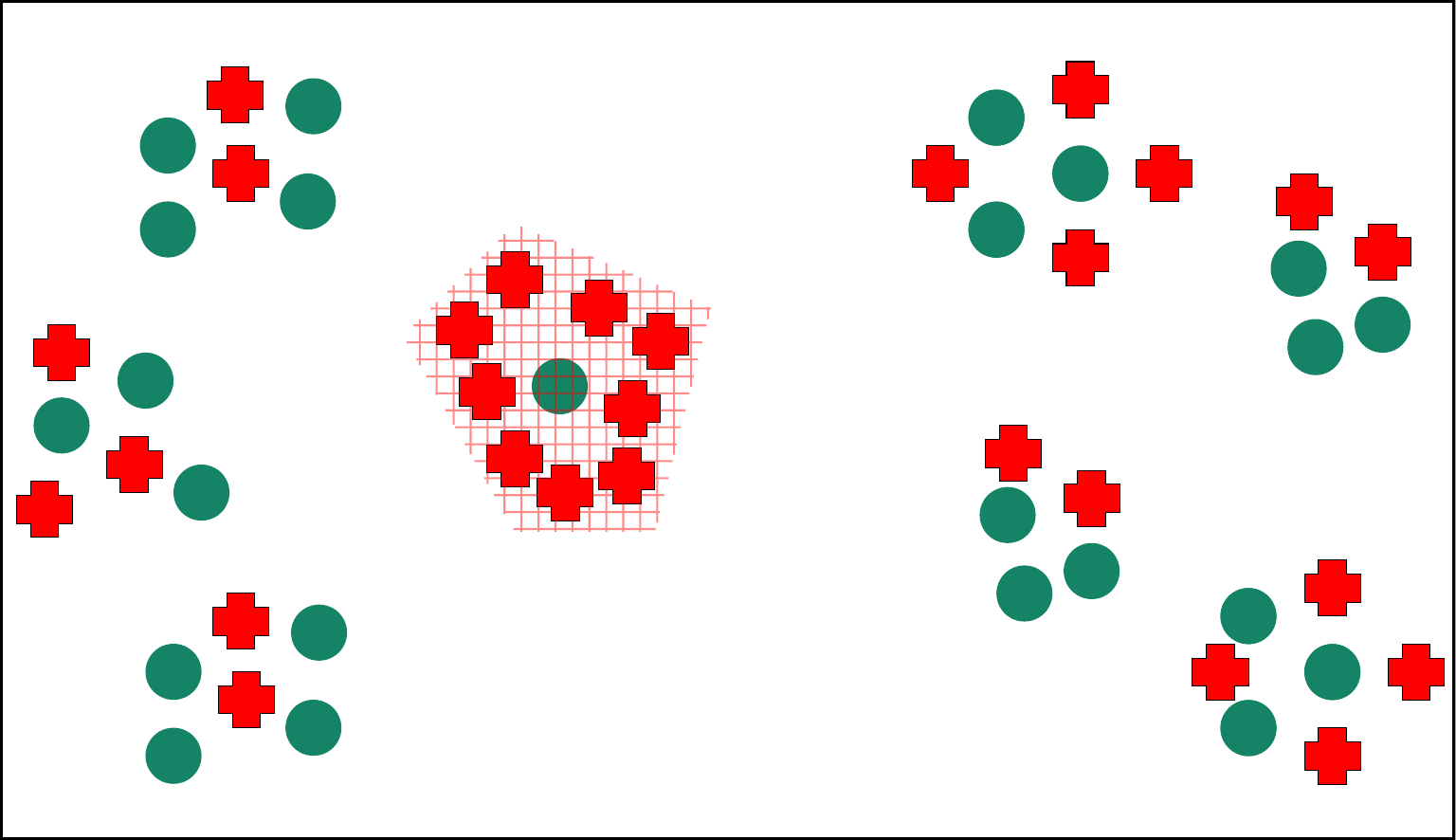}
\vspace{0.3em}
\includegraphics[width=0.45\linewidth]{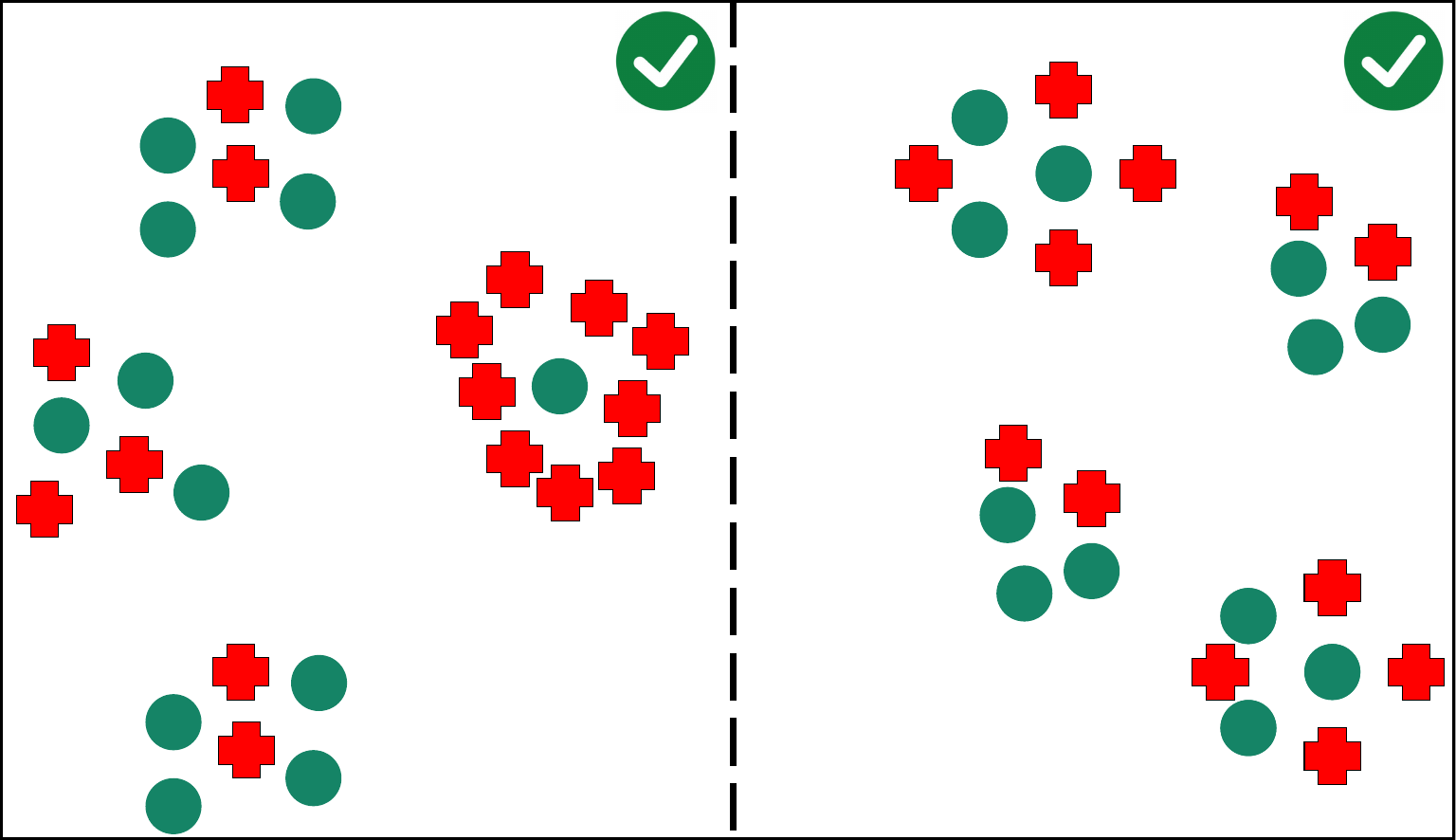}
\vspace{0.3em}

\includegraphics[width=0.45\linewidth]{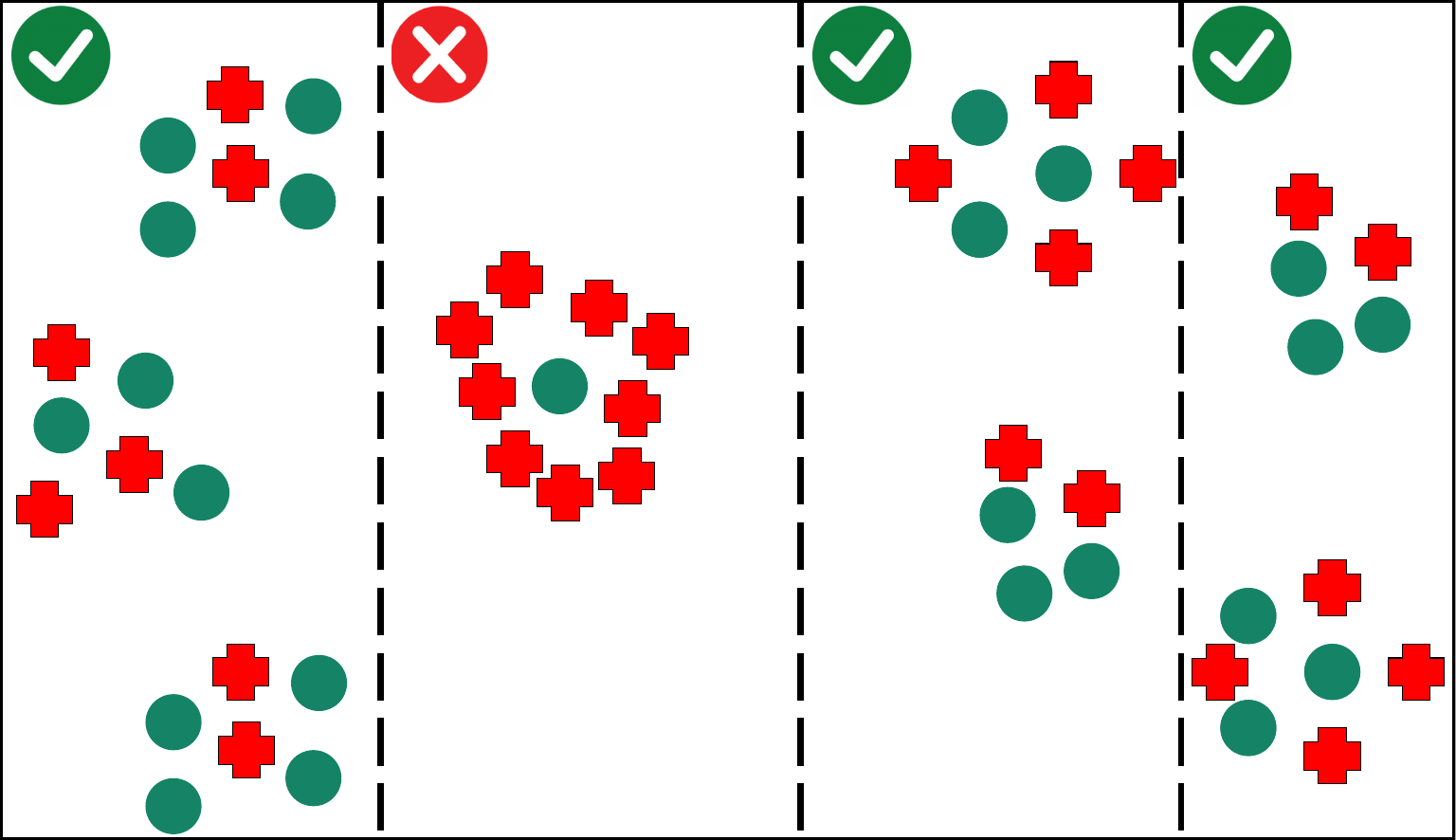}
\includegraphics[width=0.45\linewidth]{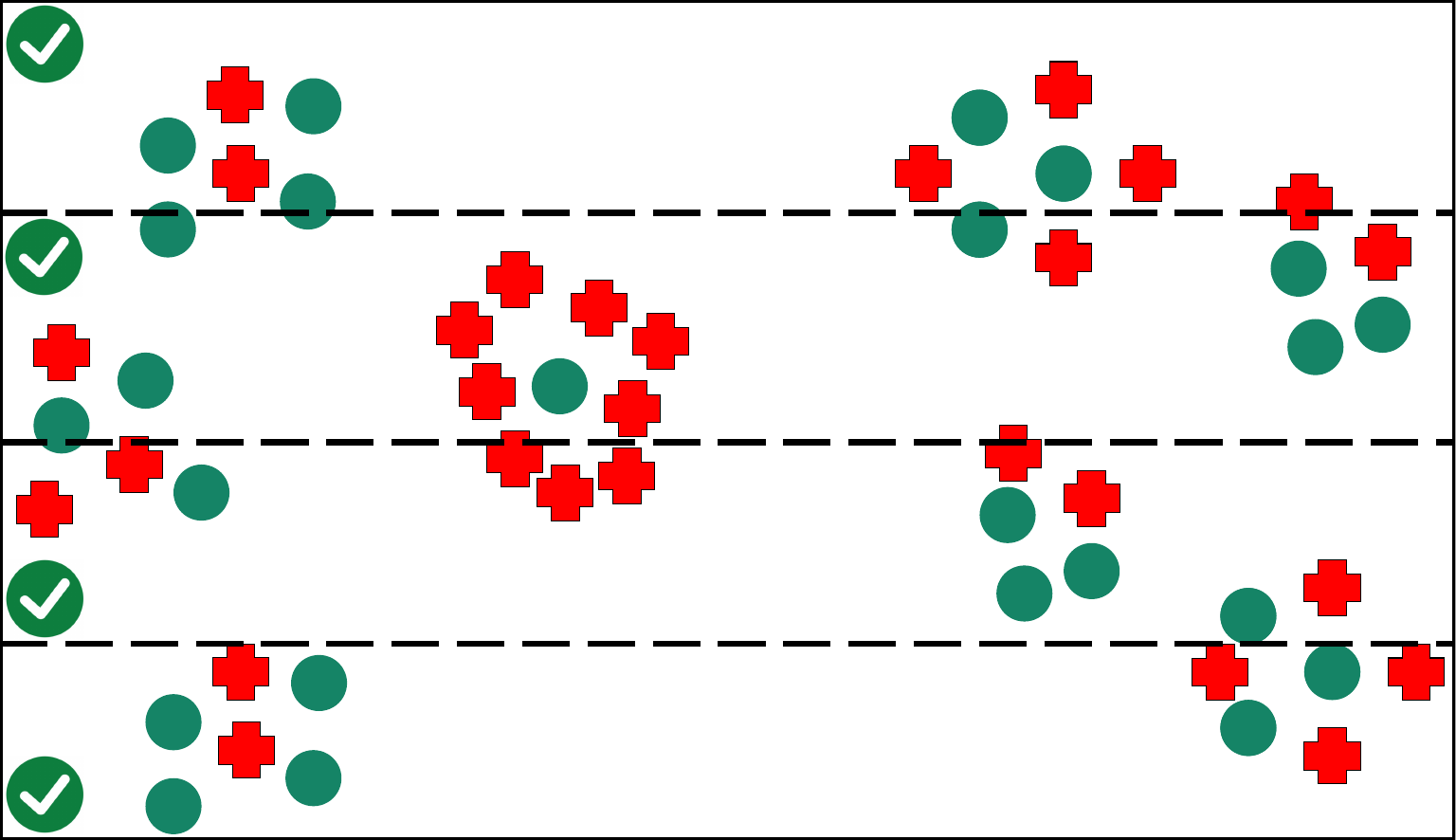}

\caption{Simple example illustrating the Modifiable Areal Unit Problem when assessing the spatial fairness of a binary classifier with respect to possible different partitionings. A positive (negative) outcome is denoted in green (red).}
\Description{Four plots illustrating the Modifiable Areal Unit Problem for spatial fairness assessment of a binary classifier. Green points represent positive outcomes and red points represent negative outcomes. In the top-left plot, a highlighted central hotspot contains predominantly negative outcomes relative to the overall population. In the top-right plot, a coarse two-cell vertical partition mixes the hotspot with surrounding observations, so no cell exhibits a sufficiently distinct outcome distribution and the hotspot is missed. In the bottom-left plot, a finer four-cell vertical partition isolates the hotspot more effectively, allowing the unfair region to be detected. In the bottom-right plot, a four-cell horizontal partition again mixes the hotspot with surrounding observations and fails to detect it. The example demonstrates that both partition resolution and alignment can determine whether localized unfairness is detected.}
\label{fig:maup}
\end{figure}

To better illustrate the Modifiable Areal Unit Problem (MAUP), consider the simple example in Figure \ref{fig:maup}, which shows the case of assessing the spatial fairness of a binary classifier. The top-left plot shows the spatial distribution of the classifier’s outcomes (positive in green, negative in red).
We consider $52$ individuals: $28$ receive a negative outcome and $24$ a positive one, yielding a global positive rate of $\sim 0.46$ and a global negative rate of $\sim 0.54$. Our goal is to determine whether these global rates also hold locally, or whether there exist areas in which the classifier behaves differently: we call these areas \textit{hotspots of unfairness}.
In practice, this requires identifying regions where the local positive/negative rates differ significantly from the global ones.

In the top-left plot of Figure \ref{fig:maup}, the area highlighted in red provides an example of a hotspot: it contains individuals receiving predominantly negative outcomes (local rates: negative $0.9$, positive $0.1$). 
Unfortunately, the shape and location of hotspots of unfairness are not known \emph{a priori}. As a result, relying on a single spatial partition would fail to detect localized instances of unfairness.

Figure \ref{fig:maup}'s top-right plot shows exactly this failure. Using a $1\times 2$ uniform grid, the cell containing the hotspot also includes many surrounding points:
this makes the cell’s local negative ($\sim 0.57$) and positive ($\sim 0.43$) rates remain close to the global ones, and the unfairness is not detected. Therefore, to mitigate the MAUP, related works typically repeat the assessment over multiple partitions with different resolutions and alignments, and then combine the evidence across partitions.

For instance, consider the $1\times 4$ grid used in the bottom-left plot. Here, one cell isolates the hotspot more effectively, producing a clear discrepancy between the local and global rates. However, varying resolution alone may still be insufficient; changing the grid alignment can be crucial as well. This is illustrated by the $4\times 1$ grid used in Figure \ref{fig:maup}'s bottom-right plot, where the local rates within each cell remain close to the global ones, and the hotspot is not detected.

Overall, this simple example shows how different partitions with varying resolutions and alignments can be used to deal with the MAUP. It also motivates the need to combine evidence gathered across multiple partitions to draw final conclusions. The same considerations apply to the problem of assessing predictive models for fairness based on activity-space patterns.

\section{Further details on the step of computing the subsets of cells that must undergo hypothesis testing}
\label{app: supplemental complexity cand gen}

The Eclat-like algorithm \citep{zaki2000scalable} we implemented to determine the subsets of cells that must undergo hypothesis testing is a (1) \textit{bottom-up} approach, in that it finds frequent subsets of cells of cardinality $k$ by joining the frequent subsets of cardinality $k-1$ found during the previous iteration; (2) the support of a subset of cells is computed as the cardinality of the intersection between the sets of moving object IDs associated with the subsets being joined; (3) it is a \textit{prefix-based} approach, meaning that each cell is assigned a unique ID, enabling lexicographic sorting of subsets of cells, i.e., the prefixes\footnote{We report that each prefix technically corresponds to an equivalence class, as introduced in the Definition 8 from \citep{zaki2000scalable}.}. 
This, in turn, allows efficient partitioning of the search space by reducing redundant joins. 
Finally, while more efficient algorithms than Eclat-like ones exist, e.g., those targeting maximal itemsets to speed up the computation of candidate itemsets \citep{han2000mining, zaki2000scalable}), an Eclat-like algorithm integrates very well with our assessment framework.

Finally, we highlight a theoretical result from \citep{zaki2000scalable} that applies to Eclat-like algorithms. Among the corollaries of Theorem 3, the author observes that if $V$ is the number of transactions and the largest transaction has length $i$, then all itemsets can be enumerated and checked for support in time $O(2^i V)$: hence, for sufficiently small values of $i$, the overall complexity grows approximately linearly with the number of transactions, $\sim O(V)$. Recall then that, during the \textit{Object-to-cells mapping} step, our approach explicitly controls the maximum size $i$ a moving object’s cellset $GZ_{mo}^i$ can have. This implies that our assessment approach directly \textit{bounds} the largest possible transaction size, and when stop segments are used as the basis for the object-to-cells mapping step (as we do in this work), $i$ can be expected to be reasonably small.

\section{Observations on the complexity of the hypothesis testing step}
\label{app: supplemental complexity hyp test}

The overall complexity of the hypothesis testing step is dominated by the $n$ Monte Carlo simulations needed to derive an approximated distribution of $T$ under $H_0$, i.e., $O(|\mathit{PRED}| \times |\mathit{CANDIDATES}_{\mathcal{GZ}}| \times n)$. More specifically, for each simulation $b \in \{1, \ldots, n\}$ we need to compute the test statistic $T$. This requires to first permute the predicted values, which has cost $O(|PRED|)$; and then compute $L_1^{max}(\mathit{CANDIDATES}_{\mathcal{GZ}}, \mathit{PRED}_b)$, which requires, for every subset $c \in CANDIDATES_{\mathcal{GZ}}$, to aggregate the predicted values of the moving objects associated with $c$ and the predicted values of the other moving objects. The cost per subset $c$ is $O(|\mathit{PRED}|)$, hence evaluating $L_1^{max}$ for all subsets $c \in \mathit{CANDIDATES}_{\mathcal{GZ}}$ has cost $O(|\mathit{CANDIDATES}_{\mathcal{GZ}}| \times |\mathit{PRED}|)$. Finally, observe that computing $L_0^{max}(PRED)$ can be done just once, and the result can be shared across the simulations. Therefore, each simulation has cost $O(|\mathit{CANDIDATES}_{\mathcal{GZ}}| \times |\mathit{PRED}|)$.

\section{Overall complexity of the proposed approach}
\label{app: supplemental complexity approach}

Let us briefly discuss the overall complexity of our assessment approach. Let $S_1$ be the trajectory segmentation step, $S_2$ the geographic zoning step, $S_3$ the object-to-cells mapping step, $S_4$ the step that computes the subsets of cells that must undergo hypothesis testing, and $S_5$ the hypothesis testing step. 
If we denote as $q$ the number of geographic zonings used, then the overall complexity of our approach is: 
$$O\Big(S_1 + S_2 + q \cdot (S_3 + S_4) + S_5 \Big).$$
In practice, we highlight that the time-dominant steps are $S_3$, $S_4$, and $S_5$. However, observe that $S_3$ and $S_4$ are embarrassingly parallel with respect to the different zonings, while $S_5$ is embarrassingly parallel with respect to the Monte Carlo simulations (i.e., the time-dominant component within that step). Consequently, increasing the computational resources allocated to these steps reduces their execution time.

\section{Understanding sensitivity and positive predictive value (PPV): a toy example}
\label{app: supplemental example sensitivity ppv}

To help the reader develop intuition on these two metrics, consider a dataset containing 100,000 moving objects, and suppose that unfairness has been injected through  their labels such that there is a hotspot of unfairness composed of two distinct regions. Assume that this hotspot is associated with 600 objects, i.e., these 600 objects are associated with its two regions and are treated differently because of this. Now suppose that our assessment approach uses a set of uniform grids (geographical zonings) with different resolutions and alignments, and returns from the various grids a collection of subsets of cells that are deemed extreme. Assume further that the total number of objects associated with any of these subsets is 1,000, and that 300 of these objects actually belong to the hotspot. Then, sensitivity is equal to 0.5 and PPV to 0.3.

Consider now a limit case, in which we artificially make the assessment approach return the entire set of 100,000 objects whenever it detects unfairness. In this case, sensitivity would be a perfect 1, but PPV would be only 0.006, indicating that localization performance is, effectively, compromised.
Consider finally a second limit case, in which the assessment approach always returns objects belonging to a hotspot, but only a small fraction of them; for instance, suppose that it returns only 30 of the hotspot’s 600 objects. In this case, the approach would achieve a perfect PPV of 1, but its sensitivity would only be 0.05, which indicates another type of poor localization performance.

In general, observe the tension between sensitivity and PPV: they capture complementary aspects of localization. High sensitivity with low PPV indicates that the detected hotspot(s) covers a large portion of the true unfair hotspot(s) but extends substantially into non-unfair areas. In contrast, high PPV with low sensitivity indicates that the detected hotspot(s) is spatially precise, but captures only a limited portion of the true unfair hotspot(s)'s extent.

\section{Hypothesis testing when assessing a binary classifier for fairness based on activity-space patterns.}
\label{app: supplemental hyp test binary class}

We first state the null and alternative hypotheses (Stage 1). Next, we select the scan statistic appropriate for assessing binary classifiers for fairness based on activity-space patterns, i.e., the Bernoulli-based spatial scan statistic from \citep{kulldorff1997spatial} (Stage 2). Under the Bernoulli model, both hypotheses are expressed through the binomial parametric probability model 
$\binom{N}{k}\theta^k \times (1 - \theta)^{N-k},$ 
where $N$ is the number of trials, $k$ the number of successes, and $\theta$ the probability of success. 

We then formalize $H_0$ as the likelihood function $L_0$ using the above-mentioned parametric probability model (stage 3):
$$L_0(\theta, \mathit{PRED}) = \theta^{p(\mathit{PRED})} \times (1 - \theta)^{n(\mathit{PRED}) - p(\mathit{PRED})}$$
Here, $n(\mathit{PRED})$ is the number of labels (moving objects) in $PRED$ and $p(PRED)$ is the number of positive labels in $PRED$. Note that the binomial coefficient is omitted from both $L_0$ and $L_1$, since it would cancel out in the likelihood ratio.
Then, computing $L_0^{max}$ requires finding the MLE $\hat{\theta}$, which in this case has closed form as it can be obtained by finding the solution for the derivative $dL_0/d\theta=0$ (observe that in reality one works with the logarithm of $L_0$ for numerical stability): this yields $\hat{\theta} = p(\mathit{PRED})/n(\mathit{PRED})$.

Next (Stage 4), we formalize $H_1$ as the likelihood function $L_1$: 
\begin{align*}
& L_1(\theta_{in}, \theta_{out}, c, \mathit{PRED}) = \\
& \Big( \theta_{in}^{p(c)} \times (1 - \theta_{in})^{n(c) - p(c)} \Big) \times \\
& \Big( \theta_{out}^{p(\mathit{PRED}) - p(c)} \times (1 - \theta_{out})^{n(\mathit{PRED}) - n(c) - (p(\mathit{PRED}) - p(c))} \Big).
\end{align*}
Observe that the binomial parametric probability model is used twice in $L_1$: the first factor (second row) models the distribution of the labels over the moving objects associated with a subset of cells $c$, while the second term (third row) models the distribution of the labels over the other moving objects.
To compute $L_1^{max}$ we then proceed as follows: first, for every $c \in \mathit{CANDIDATES}_{\mathcal{GZ}}$ we compute $L_1^{max}(c, \mathit{PRED})$, which requires to compute the MLEs $\hat{\theta_{in}^c}$ and $\hat{\theta_{out}^c}$. These, in turn, are found by finding the solution for the partial derivatives:
\begin{align*}
\frac{\partial \big(L_1(\theta_{in}^c, \theta_{out}^c, c, \mathit{PRED}) \big)}{\partial\theta_{in}} = 0 \\ \frac{\partial \big(L_1(\theta_{in}^c, \theta_{out}^c, c, \mathit{PRED}) \big)}{\partial\theta_{out}} = 0.
\end{align*}
Note that, as said previously, in practice one works with the logarithm of $L_1$.
Thus, we get $\hat{\theta_{in}^c} = p(c)/n(c)$ and $\hat{\theta_{out}^c} = \big(p(\mathit{PRED})-p(c)\big)/\big(n(\mathit{PRED})-n(c)\big)$. Finally, $L_1^{max}(\mathit{CANDIDATES}_{\mathcal{GZ}}, \mathit{PRED})$ is obtained by selecting $c^* \in \mathit{CANDIDATES}_{\mathcal{GZ}}$ that yields the largest $L_1^{max}(c, \mathit{PRED})$.

The remaining operations follow the Stages 5-9 of the testing methodology verbatim, i.e., we set the desired significance level $\alpha$ (Stage 5), then compute an approximated distribution of the maximum likelihood ratio under $H_0$ by performing $n$ Monte Carlo simulations (Stage 6), subsequently compute the maximum likelihood ratio over the original predicted values in $\mathit{PRED}$, obtaining $T_{obs}$ (Stage 7), use the approximated distribution and $T_{obs}$ to determine the Monte Carlo p-value $\hat{p}$ needed to decide whether to reject $H_0$ (Stage 8), and finally store the evidence of unfairness (if any) in $\mathit{UNFAIR}_{\mathcal{GZ}}$ (Stage 9).

\end{appendices}

\bibliographystyle{ACM-Reference-Format}
\bibliography{biblio}

\end{document}